\documentclass[
12pt, 
a4paper, 
oneside, 
BCOR=5mm, 
]{scrartcl}
\usepackage{setspace} 
 \makeatletter
\DeclareOldFontCommand{\rm}{\normalfont\rmfamily}{\mathrm}
 \DeclareOldFontCommand{\sf}{\normalfont\sffamily}{\mathsf}
 \DeclareOldFontCommand{\tt}{\normalfont\ttfamily}{\mathtt}
 \DeclareOldFontCommand{\bf}{\normalfont\bfseries}{\mathbf}
 \DeclareOldFontCommand{\it}{\normalfont\itshape}{\mathit}
 \DeclareOldFontCommand{\sl}{\normalfont\slshape}{\@nomath\sl}
 \DeclareOldFontCommand{\sc}{\normalfont\scshape}{\@nomath\sc}
 \makeatother
\usepackage[english]{babel}
\usepackage[T1]{fontenc} 
\usepackage{times}
\usepackage[utf8]{inputenc}
\usepackage{newunicodechar}
\newunicodechar{〈}{\langle}
\usepackage{tabularx}
\usepackage{graphicx} 
\graphicspath{{Figures/}} 

\usepackage{enumitem} 
\newlist{steps}{enumerate}{1}
\setlist[steps, 1]{label = Step \arabic*:}

\usepackage{lipsum} 

\usepackage{amsmath,amssymb,amsthm} 
\usepackage{varioref} 
\usepackage{booktabs} 

\usepackage{siunitx} 

\usepackage{pgfplotstable} 

\usepackage{etoolbox}

\usepackage{lipsum}

\usepackage{float}

\usepackage{amsmath}

\usepackage[T1]{fontenc}

\usepackage{natbib}

\usepackage{listings}

\usepackage[labelfont=bf]{caption}
\usepackage{longtable}

\usepackage{filecontents}  

\usepackage{amsfonts}

\usepackage[colorinlistoftodos]{todonotes}

\usepackage{algorithm,algpseudocode}
\usepackage{algorithmicx}

\usepackage{etoolbox}

\usepackage{blindtext}

\usepackage{enumitem}
\usepackage{lmodern}
\usepackage{fullpage}
\usepackage{alltt}
\usepackage{boxedminipage}
\usepackage{color}

\usepackage{booktabs}

  \usepackage{url}

\usepackage[font={normal,it}]{caption}

\theoremstyle{definition} 

\theoremstyle{plain} 

\theoremstyle{remark} 

\usepackage{authblk}
\usepackage{chngcntr}
\counterwithin{figure}{section}
\PassOptionsToPackage{hyphens}{url}
\usepackage{breakcites}
\usepackage{caption}
\usepackage{listings}

\date{} 

\let\clsCenter\Center\let\clsendCenter\endCenter
\let\Center\undefined\let\endCenter\undefined
\usepackage{ragged2e}
\let\Center\clsCenter
\let\endCenter\clsendCenter

\newcommand\SentenceCase[1]{%
  \caselower{}%
  \capitalize{\thestring}%
}
\begin{singlespace}
\title{Combining K-means type algorithms with Hill Climbing for\\
Joint Stratification and\\
Sample Allocation Designs} 
\author
{Mervyn O'Luing,$^{1}$ Steven Prestwich,$^{1}$ S. Armagan Tarim$^{2}$
}
\end{singlespace}
\usepackage{subfig}
\usepackage{bibentry}

\pgfplotsset{compat=1.15}
\begin{document}

\maketitle

\begin{singlespace}
\begin{center}
    \textbf{Abstract}
\end{center}


In this paper we combine the k-means and/or k-means type algorithms with a hill climbing algorithm in stages to solve the joint stratification and sample allocation problem. This is a combinatorial optimisation problem in which we search for the optimal stratification from the set of all possible stratifications of basic strata. Each stratification being a solution the quality of which is measured by its cost. This problem is intractable for larger sets. Furthermore evaluating the cost of each solution is expensive. A number of heuristic algorithms have already been developed to solve this problem with the aim of finding acceptable solutions in reasonable computation times. However, the heuristics for these algorithms need to be trained in order to optimise performance in each instance. We compare the above multi-stage combination of algorithms with three recent algorithms and report the solution costs, evaluation times and training times. The multi-stage combinations generally compare well with the recent algorithms both in the case of atomic and continuous strata and provide the survey designer with a greater choice of algorithms to choose from. \\



{\let\thefootnote\relax\footnotetext{\textsuperscript{1} \textit{Insight Centre for Data Analytics, Department of Computer Science, University College Cork, Ireland.} Email: {mervyn.oluing@insight-centre.org},{ steven.prestwich@insight-centre.org }}}

{\let\thefootnote\relax\footnotetext{\textsuperscript{2} \textit{Cork University Business School, University College Cork, Ireland.} Email: {armagan.tarim@ucc.ie}}}

\end{singlespace}


\section{Introduction}

In the joint stratification and sample allocation problem we search for a mutually exclusive and exhaustive stratification, from the set of all possible stratifications, of L smaller basic strata (i.e. strata constructed from categorical variables which are known as atomic strata or continuous variables known as continuous strata). 
The quality of each stratification (or solution) is measured by the its sample cost. Our goal is to arrange the number of strata, H, as well as the number of basic strata, $n_h$, within each stratum in order to best minimise cost (or maximise solution quality). 
\par
A solution with the maximum solution quality is said to be optimal (there may be more than one) and may be found by computing the sample allocation for each possible stratification. In other words, this optimal solution is unknown or latent. However, this exhaustive evaluation of each candidate solution, or grid-search, quickly becomes intractable for larger sets of possible solutions. This is further compounded by the computation time necessary to evaluate each solution. For this reason a number of heuristic algorithms have been proposed to search for solutions that are of an acceptable quality, e.g. a local minimum that is in close proximity to the global minimum, in a tolerable computing time.\par
In survey methodology, when clustering is referenced in relation to sampling it usually refers to a two-step process of firstly dividing the population into clusters, and secondly randomly selecting a small number of these in order to take measurements for every unit within. In this paper our focus will be on the first step - as it is applied in the unsupervised machine learning aspect of clustering – where similar units of a population are grouped together into clusters. 
Indeed the search for the optimal solution resembles clustering in the sense that we seek to group basic strata with similar characteristics together into common groups. Accordingly, for our purposes, clusters are equivalent to strata, and the similarity within each cluster will enable more precise sample allocations. \par
The equivalency of the stratification aspect of the joint stratification and sample allocation problem to a machine learning clustering technique, the K-means method, was first noted by \citep{lisic2018optimal}. This refers to a family of algorithms “developed as a result of independent investigations in the 1950s” \citep{perez2019k} but first given the name K-means by \citep{macqueen1967some}. The K-means algorithm separates observations into K-clusters, so that the within-cluster sum of squared errors (SSE) is minimised \citep{hartigan1979k}.  The purpose of the algorithm is not to find some unique, definitive grouping, but rather to simply aid the investigator in obtaining a qualitative and quantitative understanding of large amounts of N-dimensional data by providing him/her with reasonably good similarity groups \citep{macqueen1967some}.\par
K-means is a greedy algorithm \citep{jain2010data} that finds a local minimum for a problem that is known to be NP hard \citep{drineas2004clustering} and this has led to various heuristic approaches being used to solve the problem. K-means algorithms, e.g. \citep{macqueen1967some}, \citep{lloyd1982least}, \citep{forgy1965cluster} and \citep{hartigan1979k} remain commonly used, even though many other similar clustering techniques have been introduced since their inception, e.g. expectation-maximization (EM) \citep{dempster1977maximum}, fuzzy K-means clustering \citep{bezdek1984fcm}, self-organising maps (SOM) \citep{kohonen1990self} and Neural gas \citep{martinetz1991neural}, mainly due to their ease of implementation, computational efficiency and low memory consumption \citep{morissette2013k}.\par
K-means algorithms are local search heuristics, with a number of choice inputs, such as $K$ and the initial centroids. They are sensitive to the initial centroids chosen and while this is what makes them more likely of finding local minima (more often than global minima) it also means that they sometimes produce counter-intuitive results \citep{dirgova2016neural}. Secondly, K-means algorithms provide results that are robust in nature with the implication that the local minimum is harder to escape from and the structure of the solution (i.e. number of clusters and constituents of each cluster) may be far from that of the optimal solution.\par
Thirdly, the algorithm is also known to be sensitive to the order at which basic strata would be re-assigned to clusters \citep{morissette2013k}. Fourthly, K-means also has a tendency to create clusters of equal size even if this does not represent the best clustering of the data \citep{ayramo2006introduction}.   Nonetheless, these issues are commonly mitigated by re-starting the algorithm a number of times along with varying the number of $K$, and selecting the best quality solution. \par
Following \citep{lisic2018optimal}’s proposal  \citep{ballin2020optimization} and \citep{o2020simulated} used an efficient version of the K-means clustering algorithm \citep{hartigan1979k} to determine initial solutions for our problem which are then improved upon or "optimised" using a grouping genetic algorithm and simulated annealing algorithm respectively. \citep{hartigan1979k}’s algorithm searches for a solution that is locally optimal and although it is quick, alternative solutions to that found by the algorithm may have the same or smaller within sums of squares. \par
The proximity of the local optimum to the global optimum as well as the degree of similarity within each cluster will have implications on the search for the latter by the genetic and simulated annealing algorithms. The further the local optimum is from the global optimum, the longer it is likely to take. Furthermore the grouping genetic algorithm and simulated annealing require that their hyperparameters be adequately suited to the problem in order to operate efficiently. Training the hyperparameters can also be computationally expensive.\par
Using \citep{hartigan1979k} also 
entails that all basic strata within a radius of a centroid will be assigned to a given cluster \citep{ankitchoudhary2015}. This works well when the natural data groupings are spherical in nature and clusters are linearly separable, with a corresponding high degree of similarity within clusters low degree of similarity between clusters. Indeed it has been shown that K-means can converge to a global minimum when clusters are well-separated \citep{meilua2006uniqueness}.\par
However, it is not likely to work as well when there is not a clear separation between clusters and also where there is an occurrence of basic strata from exogenous clusters within cluster spheres, i.e.  they may be close to a particular cluster centroid, but are more suited to another cluster.\par  
In this paper we explore staged combinations of K-means, expectation maximisation, self-organising map, fuzzy clustering and neural gas algorithms with hill climbing. Our aim is to find better quality solutions in reduced training times when compared to our earlier experiments on the same data with the grouping genetic algorithm and simulated annealing algorithm in \citep{o2020simulated}.\par Chapter \ref{joint} defines the objective function. Chapter \ref{hart-wong} summarises the K-means algorithm. Chapter \ref{EM1} discusses Expectation Maximisation. Chapter \ref{SOM} talks about the self-organising map. Chapter \ref{fuzzy1} describes the fuzzy clustering algorithm. Chapter \ref{neural} outlines the Neural Gas algorithm. The hill climbing algorithm with delta evaluation is described in chapter \ref{hill}. Chapter \ref{specs} provides the details of the computer specifications. This is followed by empirical comparisons for atomic and continuous strata in chapters \ref{emp_atom} and \ref{emp_contin}. The conclusions and suggestions for future work in chapter \ref{Conclusions} closes the paper. 

\section{Objective Function} \label{joint}
 
Our objective function is defined as follows:



\begin{equation}
\begin{array}{l}
\min n = \sum_{h=1}^{H}n_{h}\\
s.t.\;\; CV\left(\hat{T}_g\right)\leq \varepsilon_g\notag \;\;\; (g=1,\ldots,G)\\
2 \leq n_{h} \leq N_{h}
\end{array}
\end{equation}

where $n$ is the sample size, $n_h$ is the sample size for stratum $h$, $N_h$ is the number of units in stratum $h$ and $H$ is the number of strata. $\hat{T}_g$ is the estimator of the population total for each one of $G$ target variable columns. The upper limit of precision $\varepsilon_g$ is expressed as a coefficient of variation $CV$ for each $\hat{T}_g$. 
We solve the allocation problem for a particular stratification with the Bethel-Chromy algorithm \citep{bethel1985optimum,bethel1989sample,chromy1987design}. 

\section{The Haritgan-Wong version of the K-means algorithm} \label{hart-wong}
As described above \citep{ballin2020optimization} and \citep{o2020simulated} use an efficient version of the K-means algorithm to generate starting solutions for the genetic and simulated annealing algorithms respectively. The \citep{hartigan1979k} version seeks to attain the locally optimal minimum sums of squares within each cluster by moving data-points from one cluster to another. This means that although a sample unit might be closer its current cluster centroid, it could still be re-assigned to another cluster, if in doing so the sums of squares was reduced. 

\begin{singlespace}
\begin{algorithm}
\caption{Haritgan Wong K-means clustering algorithm}
\label{Hartigan-Wong}
\begin{algorithmic}[1]
\State Initialise solution with $K$ clusters ($k=1,...,K$) and compute centroid, $c_k$ for each cluster $C_k$
\State For each $\bar{Y_g}$, in sample unit $l$, calculate the Euclidean distance, $dist=\sqrt{\sum_{g}^{G} (c_g - \bar{Y_g})}$, from the centroids $c_k$ of $K$ clusters $C_k$.
\State Assign basic strata to clusters with the closest centroids
\State Calculate the new centroids of the clusters $c_k=\frac{\sum_{\bar{Y}_{g\in c_k}}\bar{Y}_{g}}{\left | c_k \right |}$
\State Compute $SSE\left ( C_g \right )= \sum_{g} \sum_{\bar{Y} \in C_g} dist^2 \left (c_g , \bar{Y} \right )$
\State Repeat Step 2 to Step 5 until convergence.
\end{algorithmic}
\end{algorithm}
\end{singlespace}

The algorithm is deemed to have converged when no sample unit changes cluster in the next iteration or there is an increase in SSE in an iteration when compared to its previous one. We did not train any hyperparameters for the K-means algorithm.

\section{Expectation Maximisation}\label{EM1}

A Gaussian Mixtures Model considers whether each sample unit belongs to one of a mixture of, $G$, Gaussian probability distributions (each representing a given stratum). 
The Expectation-Maximisation, EM, algorithm is an iterative method that we can use to maximise the likelihood a sample unit belongs to each stratum. It is used in clustering problems where the optimal solution is latent. 
EM is similar to K-means in the action of grouping similar basic strata into strata. It differs from K-means in the sense that each sample unit has a probability of being in each stratum (i.e. soft clustering), whereas in K-means a sample unit in assigned to one stratum (hard clustering). EM is not as sensitive to the initial clustering as K-means and works well in practice. However, it is not guaranteed to find the maximum likelihood. 
EM works as follows: 

\begin{singlespace}
\begin{algorithm}[H]
\caption{Expectation-Maximisation clustering algorithm}
\label{EM3}
\begin{algorithmic}[1]
\State Initialise a random set of parameters $\hat\theta$.
\State At the $j_{th}$ step estimate the expected value for each latent stratum in the solution. $Q(\theta,  \hat\theta^{(j)})=E_{Z\mid X,\theta^{(j)}} [log L(\theta;X, Z)]$
\State Optimize the parameters of this distribution using maximum likelihood. Determine the new estimate $\hat\theta^{(j+1)}$ as the maximizer of $Q(\theta,  \hat\theta^{(j)})$ over $\theta$, where j represents the current step, $Z$ represent the latent stratification, $\theta$ represents the unknown parameters of the distributions and $X$ represents the data. 
   \item Repeat steps 2 and 3 until convergence. 
\end{algorithmic}
\end{algorithm}
\end{singlespace}

We used the \emph{estep} and \emph{mstep} functions from the \emph{mclust} package \citep{mclust2016} in R \citep{R2021} iteratively until they converged on a solution. We did not train any hyperparameters for EM.

\section{The Self-Organising Map, SOM}\label{SOM}

In a process that again is similar to K-means, we can use self-organising maps, SOMs, to group similar data together into units (or clusters) on a grid (or map). Indeed SOMs are also known as constrained K-means because in SOMs the centroids are fixed, whereas in K-means they vary according the changing data in each stratum. In SOMs each cluster is arranged in a rectangular or hexagonal grid. The clusters, however, are different to the strata discussed up to now. \par A SOM is an artificial neural network algorithm that takes the data as input and transcribes them into an output of lower dimensionality (usually bi-dimensional) \citep{morissette2013k}. In this respect, K-means is harder to visualise if the clusters are multi-dimensional. \par SOMs use a Hebbian competitive learning algorithm to build the map. The algorithm begins with initialising a set of weights. It then proceeds to find the nearest (winning) cluster to a randomly selected sample unit. The weights of winning clusters, and those of their neighbourhood, are updated. \citep{kohonen1990self} suggests that clusters have a large neighbourhood size to begin with, e.g. a radius of more than half the diameter of the grid. The learning rate (or rate of change of the weight vectors \citep{asan2012introduction}) decreases until algorithm converges and the neighbourhood value reduces gradually to $1$ usually as a function of the increase in iterations. The final weights are the referencing \emph{codebook} vectors between the clusters and their connected basic strata.  

\begin{singlespace}
\begin{algorithm}[H]
\caption{Self-Organising Maps}
\label{EM}
\begin{algorithmic}[1]
\State Initialise a set of weights.
\State Randomly selecting an input (sample unit.
\State Select the nearest (winning) cluster using Euclidean distance.
\State Update the weights for the winning cluster (and an iteratively reducing number of neighbours)
$ {\displaystyle W_{v}(s+1)=W_{v}(s)+\theta (u,v,s)\cdot \alpha (s)\cdot (X(t)-W_{v}(s))}$ , where s is the current iteration, $\lambda$  is the iteration limit, $t$ is the index of the sample unit in the input data set ${\mathbf{X}}$, ${{X}(t)}$ is a target sample unit, $v$ is the index of the node in the map,$\mathbf {W} _{v}$ is the current weight vector of node $v$,$u$ is the index of the best matching cluster (BMC) in the map, $\theta (u,v,s)$ is a restraint due to distance from BMC, usually called the neighbourhood function, and $\alpha (s)$ is the learning rate which decreases as the iterations progress.
\State Repeat steps 2 to 4 for a set number of iterations.
\end{algorithmic}
\end{algorithm}
\end{singlespace}

SOMs better explore the search space and are less prone to local optima than K-means \citep{baccao2005self}. SOMs are particularly useful in solving problems where there are multiple optima such as our problem. However, the number of grid clusters in the SOM can be quite large. The visualisations of such grids are more of qualitative importance than quantitative.\par  
The basic principle of our research is that similar basic strata be grouped together. In a detailed grid similar basic strata can be mapped to different grid clusters. It is therefore appropriate that similar clusters are grouped together into larger clusters.\par 
Clustering SOMs was first proposed by \citep{vesanto2000clustering} using both agglomerative and partitive clustering techniques, in particular K-means. However, they also refer to expectation maximisation and fuzzy clustering of the SOM. They also suggest using a neural gas algorithm to initially abstract the data (instead of the SOM).\par
\citep{vesanto2000clustering} found that a two-stage process of using SOM to create a prototype which are then clustered provides better quality output in terms of quantitative analysis (e.g. means, medians and ranges) and computation times when compared with direct clustering of the data. 
Their goal was not to find the optimal clustering for the data but to get good insight into the cluster structure of the data for data mining purposes. 
\par For SOM we use the \emph{Kohonen} package \citep{wehrens2018self} and we train the following hyperparameters: number of iterations, upper and lower $\alpha$ values and the neighbourhood radius. 

\section{Fuzzy K-means clustering}\label{fuzzy1}

In K-means clustering a sample unit belongs to one cluster only, whereas in fuzzy K-means it belongs to all clusters (or strata) at varying probabilities. In the fuzzy K-means algorithm (also called the fuzzy c-means algorithm) each centroid of a cluster ($c_{k}$) is the mean of all cases (basic strata) in the dataset, weighted by their degree of belonging to that cluster ($w_{k}$). Like K-means the algorithm is affected by the initial clusters and tends to find local minima \citep{morissette2013k}. Nonetheless, the algorithm is useful in scenarios where there is some degree of overlap between clusters and it would be useful to consider to what degree a sample unit belongs to various clusters. The soft clustering helps to smooth out the likelihood of the algorithm finding inferior quality local minima \citep{klawonn2004fuzzy}. Using the description provided in \citep{bezdek1984fcm} as a guide the algorithm operates as follows: 



\begin{singlespace}
\begin{algorithm}[H]
\caption{Fuzzy K-means}
\label{fuzzy}
\begin{algorithmic}[1]
\State Fix $K, m, A, \parallel j \parallel_{A}$. Choose an initial matrix $U^{(0)} \in M_{fK}$ the set of fuzzy partitions. Then at step j, $j, = 0, 1, ..., LMAX$, 
\State Compute means $\hat{v}^{(j)}, i = 1, 2, . . . , K$ using $\hat{v}_{i}= \frac{\sum_{j=1}^{N} (\hat{u}_{ij})^{m} y_{j}}
{\sum_{j=1}^{N} (\hat{u}_{ij})^{m} }; 1 \leqq i \leqq K$;
\State Compute an updated membership matrix $\hat{U}^{j+1} = [\hat{u}_{ij}^{j+1}]$ using $\hat{u}_{ij} =\left( \sum_{q=1}^{K}\left(\frac{\hat{d}_{ij}}{\hat{d}_{qj} }\right)^{\frac{2}{m-1}} \right)^{-1} $, where $\hat{d}_{ij}= \parallel y_{j} - \hat{v}_{i} \parallel_{A}$ , $v_{i} = (v_{i1},v_{i2}....,v_{in})$ = centre of cluster $K$
\State If $\parallel \hat{U}^{j+1} - \hat{U}^{j}\parallel < \epsilon $, stop. Otherwise set $\hat{U}^{j} = \hat{U}^{j+1}$ and return to step 2. 
\end{algorithmic}
\end{algorithm}
\end{singlespace}

The number of clusters is represented by $K$ (or $c$ in \citep{bezdek1984fcm}), $x_{j}$ is the $jth$ sample and $v_{i}$ is the centre of the $ith$ cluster. $\parallel \parallel_{A}$ represents the norm function. $U$ is a k by N matrix representation of the partition $Y_{i}$, where $\sum_{i=1}^{k} Y_{i} = Y$, $U$ =$[u_{ij}]$ and $u_{i}(y_{i}) = u_{ij} = \lbrace_{0; otherwise}^{1 ; y_{j} = Y_{i}} $. All clusters must contain at least one sample unit and the weighted probabilities for each sample unit belonging to each cluster sum to 1. $K$ is the number of clusters, $m$ is the weighting exponent, $A$ is the positive-definite $(n$ by $n)$ weight matrix, $LMAX$ is the maximum number of iterations. We use the fuzzy clustering method in the \emph{cmeans} function in the \emph{e1071 package} \citep{meyer2019package} in R. For this we train $m$ the weighting exponent. 
\section{Neural gas}\label{neural}

Developed by \citep{martinetz1991neural} the neural gas algorithm is inspired by the SOM algorithm using a Hebb-like learning rule along with a memory decay term to map multi-dimensional data to typically one or two dimensions. In SOMs the grid for the clusters is fixed and updates are made to the weights of winning clusters and clusters in their neighbourhood. In Neural Gas (NG), however, updates are made to clusters independently of their location on the grid. Instead the algorithm sorts the clusters according to the distance of their reference weights to the basic strata. The level of the adjustment to each weight is determined by their "rank" in this order \citep{Li2014Smile}. The winning weight takes the largest adjustment. The strength of the adjustment decreases according to a fixed schedule. Because prior knowledge of the grid structure is not required for neural gas, this in tandem with the soft nature of the clusterings may assist the neural gas find better quality clusters than the SOM. The algorithm can be summarised as follows: 

\begin{singlespace}
\begin{algorithm}[H]
\caption{Neural Gas}
\label{fuzzy3}
\begin{algorithmic}[1]
\State Assign initial values to the weights $w_{i} \in R^{n}$ and set all $C_{ij}$ to zero.
\State Select an sample unit, or vector v, from the input data
\State for each unit i determine the number $k_{i}$ of clusters $j$ with:
                                            $\parallel v - w_{j}\parallel < \parallel v - w_{i}\parallel$ 
    by, e.g., determining the sequence ($i_{0},i_{1},...,i_{N-1}$) of clusters with 
                                            $\parallel v - w_{i_{0}}\parallel < \parallel v - w_{i_{1}}\parallel < ...< \parallel v - w{i_{N-1}}$ 
\State Adapt the weights accordingly:
    $w^{new}_{i} = w^{old}_{i} + \epsilon . e^{\frac{- k_{i}}{\lambda} (v - w^{old}_{i})}$, $i = 1,...,N$.
\State If $C_{i_{0}i_{1}} = 0$, set $C_{i_{0}i_{1}} = 1$ and $t_{i_{0}i_{1}}=0$. If $C_{i_{0}i_{1}} = 1$, then $t_{i_{0}}{i_{1}}=1$.
\State Set  $t_{i_{0}j} = t_{i_{0}j} + 1 $ for all $j$ with $C_{i_{0}j}=1$
\State Set $C_{i_{0}j}=0$ for all $j$ with $C_{i_{0}j}=1$ and $t_{i_{0}j} > T$. Return to step 2.
\end{algorithmic}
\end{algorithm}
\end{singlespace}
where the step size $\epsilon$ $\in [0,1]$, $\lambda$ controls the number of clusters changing their weights, $w_i$, in each step, $C_{ij}$ are elements in the matrix $C$ which are set from $0$ to $1$ if clusters $i,j$ are similar, $t_{i,j}$ relates to an iteration within a set number of iterations, $T$. We implement the neural gas method using the \emph{cclust} function in the \emph{cclust} package \citep{dimitriadou2020package} - for which we fine tune the upper and lower values of the $\lambda$ and $\epsilon$ hyperparameters.


\section{Hill Climbing Algorithm with delta evaluation} \label{hill}

The term \emph{hill climbing} implies a maximisation problem, however the equivalent optimisation of the cost function can be applied in a downward sense also, i.e. for minimisation problems. Hill climbing is used to describe both types of problems without loss of generality \citep{michalewicz2013solve}. We apply hill climbing to search for better neighbouring solutions in our problem. In doing so we create a new solution by randomly taking a sample unit from one stratum in the current solution and moving it to another stratum. We then evaluate the quality of that new solution state and keep the change if its quality is found to be better. If not we return to the previous solution state. This process continues until there has been no further improvement in solution quality after a set number of iterations. The similarity between solutions was used by \citep{o2020simulated} to achieve gains in computation times by using delta evaluation for a simulated annealing algorithm.\par
Hill climbing is analogous with the advanced state of a simulated annealing algorithm when the tunable probability of accepting inferior solutions has decayed to a sufficient degree that the algorithm only accepts better quality solutions. We also apply delta evaluation in the hill-climbing to help speed up computation times.\par
We compare the outputs from combinations of one or more of the above algorithms with the hill climbing algorithm. As with the \citep{vesanto2000clustering} our goal is not to find the optimal solution, but to get better quality clusterings (strata) in terms of sample size than those attained by the GGA and SAA - in less computation and training times. \par
The reason for using hill climbing is our assumption that the starting solutions attained by various combinations of one or more of the above algorithms will be of a sufficient quality that the hill climbing algorithm need only converge on a local minimum (which will be in close proximity to the global minimum).\par
We have fixed the number of basic strata to be moved from one stratum to another as $q=1$ and set 1,000 as the \emph{stopping-point} for the number of iterations without an improvement in solution quality, after which each consecutive pair of solution qualities 1,000 iterations apart are rounded to 2 decimal places and compared. Therefore there will be no training of hyper-parameters required for this algorithm. We base the description of the algorithm below on the summary in \citep{lin1973effective}: 

\begin{singlespace}
\begin{algorithm}[H]
\caption{Hill Climbing Algorithm}
\label{SAA}
\begin{algorithmic}[1]
\State Initialise a starting solution, $S$
\State Attempt to find an improved solution $S'$ by randomly assigning $q=1$ basic strata from stratum $h$ to stratum $h'$.
\State If an improved solution is found using delta evaluation, i.e., $f(S')$ $<$ $f(S)$ , then replace $S$ by $S'$.  
\State If no improved solution can be found after $1,000$ iterations, $S$ is a locally optimum solution, alternatively repeat from Step 2.
\end{algorithmic}
\end{algorithm}
\end{singlespace}


\section{Computer Specifications}\label{specs} 

Table \ref{specifications} below is a reproduction of the equivalent table in \citep{o2020simulated} and provides details of the processing platform used for these experiments. Further details including including the R version, packages used and number of cores is available in the description provided with the original version of the table. 

\begin{table}[H]
\caption{\textbf{Specifications of the processing platform}}
\label{specifications}
\tiny
\centering
\begin{tabular}{|l|l|l|}
\hline
\textbf{Specification}                   & \textbf{Details}                              & \textbf{Notes} \\ \hline
\textbf{Processor}                       & AMD Ryzen 9 3950X 16-Core Processor, 3493 Mhz &                \\ \hline
\textbf{Cores}                           & 16 Core(s)                                    &                \\ \hline
\textbf{Logical Processors}              & 32 Logical Processor(s)                       & 32 cores in R  \\ \hline
\textbf{System Model}                    & X570 GAMING X                                 &                \\ \hline
\textbf{System Type}                     & x64-based PC                                  &                \\ \hline
\textbf{Installed Physical Memory (RAM)} & 16.0 GB                                       &                \\ \hline
\textbf{Total Virtual Memory}            & 35.7 GB                                       &                \\ \hline
\textbf{OS Name}                         & Microsoft Windows 10 Pro                      &                \\ \hline
\end{tabular}
\end{table}

\section{Empirical Comparisons for Atomic Strata}\label{emp_atom}

Atomic strata are constructed from the cross product of categorical auxiliary ($X$) variables. Each atomic stratum contains a number of records with values for their constituent numerical target ($Y$) variables. These are summarised by calculating the means and standard deviations for each target variable.  For more details see \citep{ballin2013joint}. We used the \emph{mlrMBO} package to train the hyperparameters for the experiments below in all cases except for grid size in the SOM which we adjusted according to dataset size.

\subsection{Swiss Municipalities}

The Swiss municipalities dataset \citep{barcaroli2014samplingstrata} has 2,896 records split into 7 regions (domains). We stratify and select a sample for each of these regions with the combined stratification for all regions being the solution and the total sample size indicating the quality of the solution. The target and auxiliary variables are outlined in table \ref{vars}. The auxiliary variables \emph{POPTOT.cat} and \emph{Hapoly.cat} are categorical versions of the continuous \emph{POPTOT} and \emph{Hapoly} variables. For further details on the target and auxiliary variables refer to \citep{ballin2020optimization}. 

\begin{table}[H]
\caption{\textbf{Summary of the Target and Auxiliary variables and number of records for the Swiss municipalities dataset}}
\label{vars}
\tiny
\centering
\begin{tabular}{|l|l|l|l|}
\hline
\textbf{Dataset}                                                              & \textbf{Target Variables}                                                                  & \textbf{Auxiliary Variables}                                                               & \textbf{\begin{tabular}[c]{@{}l@{}}Number of \\ Records\end{tabular}} \\ \hline
\textbf{Swiss Municipalities}                                                          & \begin{tabular}[c]{@{}l@{}}Surfacesbois,\\ Airbat\end{tabular}                             & POPTOT.cat , Hapoly.cat                                                                            & 2,896 \\ \hline               
\end{tabular}
\end{table}

In setting a seed of 1234 and a maximum number of clusters of 30 we obtained an initial K-means solution ranging between 8 and 20 clusters for the seven domains. For each domain the solution is broken down as follows: 
\begin{table}[H]
\centering
\tiny
\begin{tabular}{|l|l|l|l|l|l|l|l|l|}
\hline
\textbf{Domain/Region}               & 1     & 2     & 3     & 4     & 5     & 6     & 7     &        \\ \hline
\textbf{Number of Clusters (Strata)} & 8     & 20    & 13    & 17    & 17    & 15    & 14    & Total  \\ \hline
\textbf{Sample Size}                 & 15.96 & 45.15 & 28.16 & 46.18 & 45.74 & 29.31 & 36.41 & 246.90 \\ \hline
\end{tabular}
\end{table}

\begin{figure}[H]
\centering 
\includegraphics[]{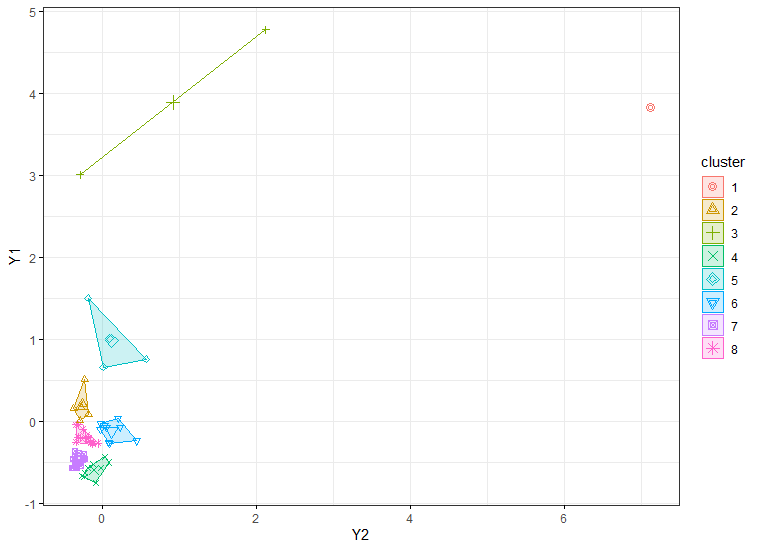}
\caption{\textbf{Cluster Plot of K-means solution represented as scaled data for Domain 1 of the Swiss Municipalities Dataset}}
\label{kmeansplot}
\end{figure}

A visual examination of the cluster plot for Domain 1 in Figure \ref{kmeansplot} indicates $8$ distinct clusters, i.e. they are linearly separated. However, the majority of clusters (i.e. those in the lower left hand corner) are not well-separated (clearly dissimilar). The sample size of $246.90$ may be further improved by considering some degree of overlap and soft clustering, i.e. whether basic strata would improve solution quality if they were in different clusters to those which they would be assigned under hard clustering. 

Table \ref{swissmunresults} below provides the sample size, execution time, total execution time and aggregated total time for fine tuning the hyperparameters for each algorithm (or staged combination of algorithms). The \textbf{Stage} variable indicates whether the sample size and times were obtained for the algorithm in one stage or whether the results were obtained through a staged combination of various algorithms. For example the sample size of $246.90$ was obtained after 1 stage by the \emph{KmeansSolution} algorithm \citep{ballin2020optimization}. This corresponds to the largest sample size. The time of $1.80$ seconds was also the quickest total execution time.\\
On the other hand the smallest sample size of $130.92$ was obtained after combining three stages of the SOM (stage 1), EM (stage 2) and Hill Climbing (stage 3) algorithms. This was achieved in a time of $178.76$ seconds ($109.84 + 68.92$). The longest time was $185.97$ seconds for the sample size of $141.46$ obtained through the staged combination of the self-organising map (SOM), K-means and hill climbing algorithms. The smallest sample size of $130.85$ should be considered in the context of the sample size of $125.17$ obtained in $248.91$ seconds after a fine-tuning time of $8,808.63$ seconds by the Simulated Annealing Algorithm (table \ref{results}). \\
Depending on the initial quality of the solution the hill climbing algorithm converges on diverse local minima resulting in varying final solution quality. Furthermore, the hill climbing algorithm search for local minima is stochastic in nature - meaning that for the same starting solution the hill climbing algorithm may also find diverse local minima on each attempt.\\
Notwithstanding the larger differences in final solution quality the same point could be made for training and the multi-stage combination of the above algorithms (or other clustering algorithms) with hill climbing. These results should not be taken as deterministic, rather as a means of providing greater scope for selecting a solution that is "good enough" in a computing time that is "small enough" \citep{sorensen2013metaheuristics}.

\begin{table}[H]
\tiny
\centering
\caption{\textbf{Comparison of Sample Size, Time and Total Execution Time for the various single algorithm or multi-algorithm combinations for the Swiss Municipalities Dataset experiment}}
\label{swissmunresults}
\begin{tabular}{|l|l|l|l|l|l|}
\hline
\textbf{Stage} &
  \textbf{Algorithm} &
  \textbf{Sample Size} &
  \textbf{Time (Seconds)} &
  \textbf{\begin{tabular}[c]{@{}l@{}}Total Execution \\ Time (Seconds)\end{tabular}} &
  \textbf{\begin{tabular}[c]{@{}l@{}}Aggregated Total \\ Time (seconds)\end{tabular}} \\ \hline
\textbf{1} &
  KmeansSolution &
  246.90 &
  1.80 &
  1.80 &
   \\ \hline
\textbf{2} &
  Hill Climbing &
  165.05 &
  33.17 &
  33.17 &
  34.97 \\ \hline
\textbf{1 and 2} &
  SOM and Kmeans &
  198.03 &
  5.55 &
  116.72 &
   \\ \hline
\textbf{3} &
  HillClimbing &
  141.46 &
  69.25 &
  69.25 &
  {\color[HTML]{F56B00} \textbf{185.97}} \\ \hline
\textbf{1 and 2} &
  NG and Kmeans &
  204.00 &
  3.34 &
  65.30 &
   \\ \hline
\textbf{3} &
  HillClimbing &
  146.25 &
  58.37 &
  58.37 &
  123.67 \\ \hline
\textbf{1} &
  \begin{tabular}[c]{@{}l@{}}Expectation \\ Maximisation\end{tabular} &
  196.12 &
  1.34 &
  1.34 &
   \\ \hline
\textbf{2} &
  HillClimbing &
  133.21 &
  68.98 &
  68.98 &
  70.32 \\ \hline
\textbf{1 and 2} &
  SOM and EM &
  183.13 &
  5.39 &
  109.84 &
   \\ \hline
\textbf{3} &
  HillClimbing &
  {\color[HTML]{32CB00} \textbf{130.92}} &
  68.92 &
  68.92 &
  178.76 \\ \hline
\textbf{1 and 2} &
  NG and EM &
  178.19 &
  1.47 &
  28.42 &
   \\ \hline
\textbf{3} &
  HillClimbing &
  134.84 &
  79.07 &
  79.07 &
  107.49 \\ \hline
\textbf{1} &
  Fuzzy Clustering &
  175.38 &
  0.27 &
  4.89 &
   \\ \hline
\textbf{2} &
  HillClimbing &
  133.91 &
  91.31 &
  91.31 &
  96.20 \\ \hline
\textbf{1 and 2} &
  SOM and FC &
  170.64 &
  3.33 &
  64.92 &
   \\ \hline
\textbf{3} &
  HillClimbing &
  139.87 &
  43.47 &
  43.47 &
  108.39 \\ \hline
\textbf{1 and 2} &
  NG and FC &
  174.34 &
  0.66 &
  13.63 &
   \\ \hline
\textbf{3} &
  HillClimbing &
  136.71 &
  71.76 &
  71.76 &
  85.39 \\ \hline
\end{tabular}
\end{table}

\begin{figure}[H]
\centering 
\includegraphics[]{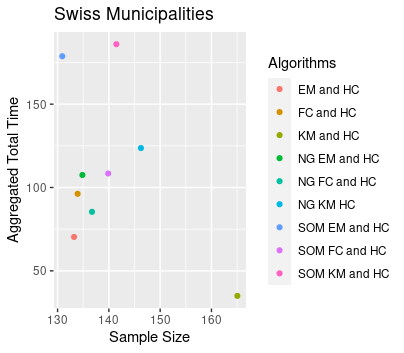}
\caption{\textbf{Plot of Aggregated Total Time versus Sample Size by Algorithm combination for the Swiss Municipalities Dataset}}
\label{AtomicSwissMunSampleTimes}
\end{figure}

Figure \ref{AtomicSwissMunSampleTimes} plots the aggregated total time against sample size by algorithm combination for the Swiss Municipalities dataset. The graph demonstrates that the SOM, EM and hill climbing combination attains the lowest sample size but with the second largest time. On the other hand, the EM and hill climbing combination attained a similar sample size but in less time. The K-means and hill climbing combination was the fastest, but the sample size was of the least quality (highest value). A number of combinations are clustered around similar sample sizes and times. 

Table \ref{SwissmunHyper} summarises the fine-tuned hyper-parameters for the algorithms where fine-tuning was implemented. 

\begin{table}[H]
\caption{\textbf{Summary of the fine-tuned hyperparameters for the various single-stage or multi-stage algorithm combinations for the Swiss Municipalities Dataset}}
\tiny
\label{SwissmunHyper}
\begin{tabular}{|l|l|l|l|l|}
\hline
                          & \textbf{Iterations}  & \textbf{Alpha High} & \textbf{Alpha Low}     & \textbf{Radius}       \\ \hline
\textbf{SOM and K-means}   & 1081                 & 0.809753218616724   & 0.031088352131124      & 0.83249740019602      \\ \hline
                          & \textbf{Lambda High} & \textbf{Lambda Low} & \textbf{Stepsize High} & \textbf{Stepsize Low} \\ \hline
\textbf{NG and K-means} & 10.6110575335055 & 0.959806751475908 & 0.499173569748991 & 0.016996916064339 \\ \hline
                          & \textbf{Iterations}  & \textbf{Alpha High} & \textbf{Alpha Low}     & \textbf{Radius}       \\ \hline
\textbf{SOM and EM}       & 7695                 & 0.113771222653394   & 0.04425198940754       & 0.061954274246033     \\ \hline
                          & \textbf{Lambda High} & \textbf{Lambda Low} & \textbf{Stepsize High} & \textbf{Stepsize Low} \\ \hline
\textbf{NG and EM}     & 8.23248667684384 & 0.67836881950614  & 0.146985992956907 & 0.032059727899148 \\ \hline
                          & \textbf{m}           & \textbf{}           & \textbf{}              & \textbf{}             \\ \hline
\textbf{Fuzzy Clustering} & 3                    &                     &                        &                       \\ \hline
                          & \textbf{Iterations}  & \textbf{Alpha High} & \textbf{Alpha Low}     & \textbf{Radius}       \\ \hline
\textbf{SOM and Fuzzy}    & 1462                 & 0.148577394044338   & 0.030700312647367      & 0.008935026169402     \\ \hline
                          & \textbf{Lambda High} & \textbf{Lambda Low} & \textbf{Stepsize High} & \textbf{Stepsize Low} \\ \hline
\textbf{NG and Fuzzy}  & 8.23248667684384 & 0.67836881950614  & 0.146985992956907 & 0.032059727899148 \\ \hline
\end{tabular}
\end{table}

\begin{figure}[H]
\centering
\includegraphics{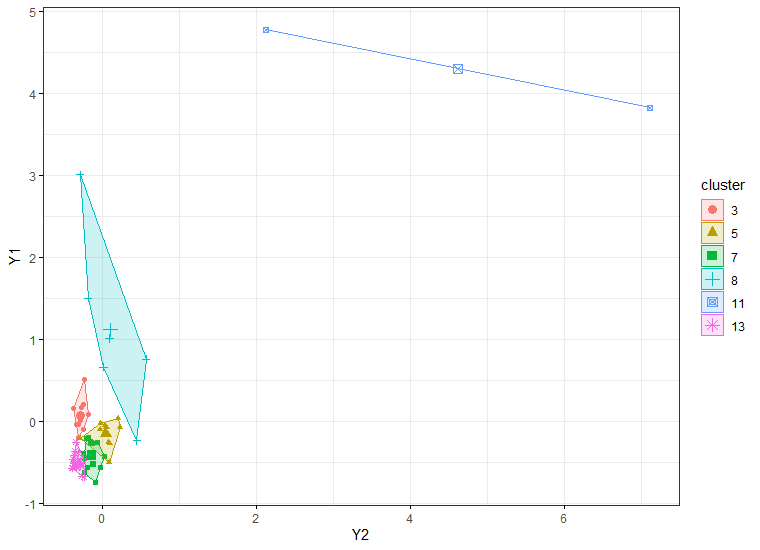}
\caption{\textbf{Cluster Plot of solution after combining SOM, EM and Hill Climbing represented as scaled data for Domain 1 of the Swiss Municipalities Dataset}}
\label{somEmHillplot}
\end{figure}

Figure \ref{somEmHillplot} visualises the clusters (strata) for Domain 1 of the Swiss Municipalities Dataset for the solution found after combining SOM, EM and hill climbing when represented as scaled data. This solution has $6$ clusters in contrast to Figure \ref{kmeansplot} which has $8$ clusters for the K-means solution. Clusters 5 and 7 show evidence of the overlap that occurs with soft clustering. Cluster 8 stretches to take in an outlier, which is unlikely to have occurred with hard clustering, or without the hill climbing algorithm.  

\subsection{American Community Survey, 2015}

A summary of the Target and Auxiliary variables and number of records for the American Survey 2015 dataset is provided in table \ref{ACSvariables} below. For further details on the dataset refer to \citep{oluing2019grouping}. 

\begin{table}[H]
\caption{\textbf{Summary of the Target and Auxiliary variables, number of records and number of domains for the American Survey 2015 dataset}}
\label{ACSvariables}
\tiny
\centering
\begin{tabular}{|l|l|l|l|l|}
\hline
\textbf{Dataset}                                                              & \textbf{Target Variables}                                                                  & \textbf{Auxiliary Variables}                                                               & \textbf{\begin{tabular}[c]{@{}l@{}}Number of \\ Records\end{tabular}}
& \textbf{\begin{tabular}[c]{@{}l@{}}Number of \\ Domains\end{tabular}}\\
 \hline
\textbf{\begin{tabular}[c]{@{}l@{}}American Community \\ Survey, 2015\end{tabular}}    & \begin{tabular}[c]{@{}l@{}}HINCP, VALP, \\ SMOCP, INSP\end{tabular}                        & \begin{tabular}[c]{@{}l@{}}BLD, TEN, WKEXREL, \\ WORKSTAT, HFL, YBL\end{tabular}           & 619,747  & 51                                                              \\ \hline
\end{tabular}
\end{table}

The sample size, execution time and total execution time for fine tuning the hyperparameters for each algorithm (or staged combination of algorithms) are outlined in table \ref{ACS} below. The sample size of 9,065.45 in a time of 8,426.67 seconds for the self-organising maps, fuzzy clustering and hill climbing combination compares favourably with the lowest sample size of 10,136.50 and a total training time of 182,152.46 seconds attained by the grouping genetic algorithm in table \ref{results}.

\begin{table}[H]
\caption{\textbf{Comparison of Sample Size, Time and Total Execution Time for the various single stage or multi-stage algorithm combinations for the American Community Survey, 2015 Dataset}}
\label{ACS}
\tiny
\centering
\begin{tabular}{|l|l|l|l|l|l|}
\hline
\textbf{Stage} &
  \textbf{Algorithm} &
  \textbf{Sample Size} &
  \textbf{Time (Seconds)} &
  \textbf{\begin{tabular}[c]{@{}l@{}}Total Execution \\ Time (Seconds)\end{tabular}} &
  \textbf{\begin{tabular}[c]{@{}l@{}}Total Training \\ Time (seconds)\end{tabular}} \\ \hline
\textbf{1} &
  KmeansSolution &
  11,769.53 &
  36.95 &
  36.95 &
   \\ \hline
\textbf{2} &
  HillClimbing &
  9,322.07 &
  5,652.85 &
  5,652.85 &
  \textbf{5,689.80} \\ \hline
\textbf{1 and 2} &
  SOM and K-means &
  12,787.19 &
  86.48 &
  2,142.86 &
   \\ \hline
\textbf{3} &
  HillClimbing &
  9,232.31 &
  6,590.34 &
  6,590.34 &
  \textbf{8,733.20} \\ \hline
\textbf{1 and 2} &
  NG and K-means &
  12,398.84 &
  107.58 &
  2,181.06 &
   \\ \hline
\textbf{3} &
  HillClimbing &
  9,290.53 &
  6,019.93 &
  6,019.93 &
  \textbf{8,200.99} \\ \hline
\textbf{1} &
  \begin{tabular}[c]{@{}l@{}}Expectation \\ Maximisation\end{tabular} &
  13,136.49 &
  32.94 &
  32.94 &
   \\ \hline
\textbf{2} &
  HillClimbing &
  9,755.61 &
  5,641.87 &
  5,641.87 &
  \textbf{5,674.81} \\ \hline
\textbf{1 and 2} &
  SOM and EM &
  13,018.89 &
  84.94 &
  12,231.26 &
   \\ \hline
\textbf{3} &
  HillClimbing &
  9,592.40 &
  6,777.39 &
  6,777.39 &
  {\color[HTML]{FE0000} \textbf{19,008.65}} \\ \hline
\textbf{1 and 2} &
  NG and EM &
  12,050.93 &
  133.64 &
  2,636.87 &
   \\ \hline
\textbf{3} &
  HillClimbing &
  9,368.93 &
  5,755.20 &
  5,755.20 &
  \textbf{8,392.07} \\ \hline
\textbf{1} &
  Fuzzy Clustering &
  13,350.35 &
  4.89 &
  352.46 &
   \\ \hline
\textbf{2} &
  HillClimbing &
  9,255.64 &
  6,245.89 &
  6,245.89 &
  \textbf{6,598.35} \\ \hline
\textbf{1 and 2} &
  SOM and FC &
  12,088.80 &
  88.87 &
  2,127.86 &
   \\ \hline
\textbf{3} &
  HillClimbing &
  {\color[HTML]{32CB00} \textbf{9,065.45}} &
  6,298.81 &
  6,298.81 &
  \textbf{8,426.67} \\ \hline
\textbf{1 and 2} &
  NG and FC &
  12,580.33 &
  123.31 &
  2,570.40 &
   \\ \hline
\textbf{3} &
  HillClimbing &
  9,249.05 &
  5,506.89 &
  5,506.89 &
  \textbf{8,077.29} \\ \hline
\end{tabular}
\end{table}

\begin{figure}[H]
\centering 
\includegraphics[]{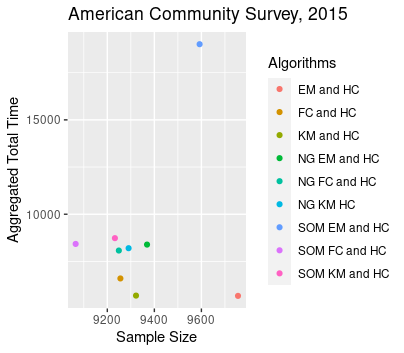}
\caption{\textbf{Plot of Aggregated Total Time versus Sample Size by Algorithm combination for the American Community Survey, 2015 Dataset}}
\label{AtomicACSSampleTimes}
\end{figure}

Figure \ref{AtomicACSSampleTimes} demonstrates that the SOM, FC and hill climbing combination attains the lowest sample size and with a moderate time. On the other hand, the EM and hill climbing combination attained the largest sample size but in the least time. A number of combinations are clustered around similar sample sizes and times. 

Table \ref{ACSfinetuning} provides the summary of the fine-tuned hyper-parameters for the relevant algorithms.

\begin{table}[H]
\caption{\textbf{Summary of the fine-tuned hyperparameters for the various single-stage or multi-stage algorithm combinations for the American Community Survey, 2015 Dataset}}
\label{ACSfinetuning}
\tiny
\centering
\begin{tabular}{|l|l|l|l|l|}
\hline
\textbf{SOM and K-means}   & \textbf{Iterations} & \textbf{Alpha High} & \textbf{Alpha Low} & \textbf{Radius}   \\ \hline
                          & 1490                & 0.211798547655744   & 0.014647548302976  & 0.569154786943982 \\ \hline
\textbf{NG and K-means} & \textbf{Lambda High} & \textbf{Lambda Low} & \textbf{Stepsize High} & \textbf{Stepsize Low} \\ \hline
                          & 27.9195691507683    & 0.953798242114481   & 0.22725191488931   & 0.005327383472354 \\ \hline
\textbf{SOM and EM}       & \textbf{Iterations} & \textbf{Alpha High} & \textbf{Alpha Low} & \textbf{Radius}   \\ \hline
                          & 351                 & 0.165795508368377   & 0.003116705920671  & 0.272235491431915 \\ \hline
\textbf{NG and EM}     & \textbf{Lambda High} & \textbf{Lambda Low} & \textbf{Stepsize High} & \textbf{Stepsize Low} \\ \hline
                          & 6.40801811285783    & 0.410143870242871   & 0.11268728621304   & 0.005130624421989 \\ \hline
\textbf{Fuzzy Clustering} & \textbf{m}          &                     &                    &                   \\ \hline
                          & 7                   &                     &                    &                   \\ \hline
\textbf{SOM and FC}    & \textbf{Iterations} & \textbf{Alpha High} & \textbf{Alpha Low} & \textbf{Radius}   \\ \hline
                          & 351                 & 0.134122729571279   & 0.002215617049982  & 0.552123088934338 \\ \hline
\textbf{NG and FC}  & \textbf{Lambda High} & \textbf{Lambda Low} & \textbf{Stepsize High} & \textbf{Stepsize Low} \\ \hline
                          & 13.0783552149509    & 0.329139476246046   & 0.384919628095097  & 0.005166985979477 \\ \hline
\end{tabular}
\end{table}

\subsection{Kiva Loans}

A summary of the target and auxiliary variables, number of records and number of domains for the Kiva Loans dataset is provided in table \ref{kivavariables} below.

\begin{table}[H]
\caption{\textbf{Summary of the Target and Auxiliary variables, number of records and number of domains for the Kiva Loans dataset}}
\label{kivavariables}
\tiny
\centering
\begin{tabular}{|l|l|l|l|l|}
\hline
\textbf{Dataset}                                                              & \textbf{Target Variables}                                                                  & \textbf{Auxiliary Variables}                                                               & \textbf{\begin{tabular}[c]{@{}l@{}}Number of \\ Records\end{tabular}} & \textbf{Number of Domains}  \\ \hline
\textbf{Kiva Loans}                                                                    & \begin{tabular}[c]{@{}l@{}}term\_in\_months,\\ lender\_count, \\ loan\_amount\end{tabular} & \begin{tabular}[c]{@{}l@{}}sector, currency, activity,\\  region, partner\_id\end{tabular} & 614,361      & 73                                                        \\ \hline
\end{tabular}
\end{table}

The sample size, execution time and total execution times for fine tuning the hyperparameters for each algorithm (or staged combination of algorithms) are outlined in table \ref{Kivaloans} below. The lowest sample size of 6,326.35 and total training time of 6,390.36 seconds for the fuzzy clustering and hill climbing combination compares favourably with the lowest sample size of 6,646.67 and total training time of 7,549.87 seconds attained by the simulated annealing algorithm in table \ref{results}.

\begin{table}[H]
\caption{\textbf{Comparison of Sample Size, Time and Total Execution Time for the various single stage or multi-stage algorithm combinations for the Kiva Loans Dataset}}
\label{Kivaloans}
\tiny
\centering
\begin{tabular}{|l|l|l|l|l|l|}
\hline
\textbf{Stage} &
  \textbf{Algorithm} &
  \textbf{Sample Size} &
  \textbf{Time (Seconds)} &
  \textbf{\begin{tabular}[c]{@{}l@{}}Total Execution \\ Time (Seconds)\end{tabular}} &
  \textbf{\begin{tabular}[c]{@{}l@{}}Total Training \\ Time (seconds)\end{tabular}} \\ \hline
\textbf{1}       & KmeansSolution           & 7,609.97                                 & 32.77     & 32.77     &                                           \\ \hline
\textbf{2}       & HillClimbing             & 6,355.91                                 & 5,088.25  & 5,088.25  & 5,121.02                                  \\ \hline
\textbf{1 and 2} & SOM and K-means           & 8,235.49                                 & 46.69     & 1,376.00  &                                           \\ \hline
\textbf{3}       & HillClimbing             & 6,686.45                                 & 4,594.16  & 4,594.16  & 5,970.16                                  \\ \hline
\textbf{1 and 2} & NG and K-means            & 7,521.78                                 & 68.12     & 1,368.16  &                                           \\ \hline
\textbf{3}       & HillClimbing             & 6,352.77                                 & 5,379.39  & 5,379.39  & 6,747.55                                  \\ \hline
\textbf{1}       & Expectation Maximisation & 8,016.05                                 & 34.89     & 34.89     &                                           \\ \hline
\textbf{2}       & HillClimbing             & 6,696.26                                 & 9,826.94  & 9,826.94  & 9,861.83                                  \\ \hline
\textbf{1 and 2} & SOM and EM               & 9,040.16                                 & 622.99    & 11,717.33 &                                           \\ \hline
\textbf{3}       & HillClimbing             & 7,173.30                                 & 12,342.69 & 12,342.69 & {\color[HTML]{FE0000} \textbf{24,060.02}} \\ \hline
\textbf{1 and 2} & NG and EM                & 7,433.73                                 & 92.04     & 1,861.40  &                                           \\ \hline
\textbf{3}       & HillClimbing             & 6,378.58                                 & 5,139.70  & 5,139.70  & 7,001.10                                  \\ \hline
\textbf{1}       & Fuzzy Clustering         & 7,532.95                                 & 17.81     & 371.75    &                                           \\ \hline
\textbf{2}       & HillClimbing             & {\color[HTML]{32CB00} \textbf{6,326.35}} & 6,018.61  & 6,018.61  & 6,390.36                                  \\ \hline
\textbf{1 and 2} & SOM and FC               & 7,707.41                                 & 50.03     & 1,357.19  &                                           \\ \hline
\textbf{3}       & HillClimbing             & 6,487.29                                 & 5,818.46  & 5,818.46  & 7,175.65                                  \\ \hline
\textbf{1 and 2} & NG and FC                & 7,470.76                                 & 84.57     & 1,671.55  &                                           \\ \hline
\textbf{3}       & HillClimbing             & 6,361.31                                 & 5,896.17  & 5,896.17  & 7,567.72                                  \\ \hline
\end{tabular}
\end{table}

\begin{figure}[H]
\centering 
\includegraphics[]{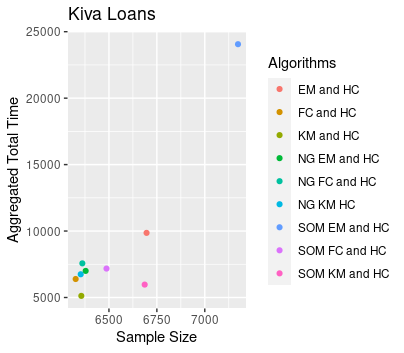}
\caption{\textbf{Plot of Aggregated Total Time versus Sample Size by Algorithm combination for the Kiva Loans Dataset}}
\label{AtomicKivaLoansSampleTimes}
\end{figure}

Figure \ref{AtomicKivaLoansSampleTimes} demonstrates that a number of combinations are clustered around similar low sample sizes and times. The FC and hill climbing combination attained the lowest sample size and with one of the lowest times. On the other hand, the SOM, EM and hill climbing combination attained the largest sample size in the most time. 

Table \ref{Kivafinetuning} provides the summary of the fine-tuned hyper-parameters for the relevant algorithms.

\begin{table}[H]
\caption{\textbf{Summary of the fine-tuned hyperparameters for the various single-stage or multi-stage algorithm combinations for the American Community Survey, 2015 Dataset}}
\label{Kivafinetuning}
\tiny
\centering
\begin{tabular}{|l|l|l|l|l|}
\hline
\textbf{SOM and K-means} & \textbf{Iterations}  & \textbf{Alpha High} & \textbf{Alpha Low}     & \textbf{Radius}       \\ \hline
\textbf{}                 & 101                 & 0.92384807669335    & 0.097029295450863  & 0.680610206832993 \\ \hline
\textbf{NG and K-means}  & \textbf{Lambda High} & \textbf{Lambda Low} & \textbf{Stepsize High} & \textbf{Stepsize Low} \\ \hline
\textbf{}                 & 16.2579368586406    & 0.068847852889988   & 0.258579228835623  & 0.005948537735878 \\ \hline
\textbf{SOM and EM}       & \textbf{Iterations} & \textbf{Alpha High} & \textbf{Alpha Low} & \textbf{Radius}   \\ \hline
\textbf{}                 & 9864                & 0.878176548467018   & 0.033124931246478  & 0.462181919813156 \\ \hline
\textbf{NG and EM}      & \textbf{Lambda High} & \textbf{Lambda Low} & \textbf{Stepsize High} & \textbf{Stepsize Low} \\ \hline
\textbf{}                 & 15.6835266864393    & 0.906532369809407   & 0.341722599971882  & 0.041738065618349 \\ \hline
\textbf{Fuzzy Clustering} & \textbf{m}          &                     &                    &                   \\ \hline
\textbf{}                 & 2                   &                     &                    &                   \\ \hline
\textbf{SOM and FC}       & \textbf{Iterations} & \textbf{Alpha High} & \textbf{Alpha Low} & \textbf{Radius}   \\ \hline
\textbf{}                 & 325                 & 0.178468058524982   & 0.015586729150577  & 0.00040918374223  \\ \hline
\textbf{NG and FC}        & \textbf{Lambda High}         & \textbf{Lambda Low}          & \textbf{Stepsize High}      & \textbf{Stepsize Low}      \\ \hline
                          & 20.0046674131416    & 0.865994203313719   & 0.23247297228314   & 0.04890728238225  \\ \hline
\end{tabular}
\end{table}

\subsection{Benchmark Results - Atomic Strata}

Table \ref{results} is a reproduction of the summary by the above datasets of the evaluation time, sample size and number of strata for the Grouping Genetic Algorithm (GGA) and Simulated Annealing Algorithm (SAA) as found by the experiments conducted by \citep{o2020simulated}. We compare these results with the above results to provide a benchmark for comparing solution quality and total training time.   

\begin{table}[H]
\caption{\textbf{Summary by dataset of the evaluation time, sample size and number of strata for the Grouping Genetic Algorithm (GGA) and Simulated Annealing Algorithm (SAA)}}
\label{results}
\centering
\tiny
\begin{tabular}{|l|l|l|l|l|l|l|}
\hline
\textbf{}                                                                           & \multicolumn{3}{l|}{\textbf{GGA}} & \multicolumn{3}{l|}{\textbf{SAA}} \\ \hline
\textbf{Dataset} &
  \textbf{\begin{tabular}[c]{@{}l@{}}Sample\\ Size\end{tabular}} &
  \textbf{\begin{tabular}[c]{@{}l@{}}Execution \\ Time\end{tabular}} &
  \textbf{\begin{tabular}[c]{@{}l@{}}Total Training \\ Time\end{tabular}} &
  \textbf{\begin{tabular}[c]{@{}l@{}}Sample \\ Size\end{tabular}} &
  \textbf{\begin{tabular}[c]{@{}l@{}}Execution \\ Time\end{tabular}} &
  \textbf{\begin{tabular}[c]{@{}l@{}}Total Training \\ Time\end{tabular}} \\ \hline
\textbf{Swiss Municipalities}               & 128.69    & 753.82       & 10,434.30  & 125.17    & 248.91       & 8,808.63  \\ \hline
\textbf{American Community Survey, 2015}         & 10,136.50 & 13,146.25    & 182,152.46 & 10,279.44 & 517.76       & 6,822.42  \\ \hline
\textbf{Kiva Loans}                         & 6,756.19  & 15,669.11    & 288,946.79 & 6,646.67  & 664.30       & 7,549.87  \\ \hline
\end{tabular}
\end{table}

\section{Empirical Comparisons for Continuous Strata}\label{emp_contin}

Unlike the atomic strata above, continuous strata are not numerical or discrete.  We use the same methodology and algorithms outlined above to the continuous strata and the results are presented below (i.e. we apply the same procedures on the continuous strata). For each of the experiments below the target and auxiliary variables are the same. 

\subsection{Swiss Municipalities}

The target and auxiliary variables for the Swiss Municipalities dataset are recorded in table \ref{SwissContinuousVars}.

\begin{table}[H]
\caption{\textbf{Summary of the Target and Auxiliary variables for the Swiss Municipalities dataset}}
\label{SwissContinuousVars}
\tiny
\centering
\begin{tabular}{|l|l|l|l|}
\hline
\textbf{Dataset}                                                              & \textbf{Target Variables}                                                                  & \textbf{Auxiliary Variables}                                                               \\ \hline
\textbf{Swiss Municipalities}                                                          & \begin{tabular}[c]{@{}l@{}}Surfacesbois,\\ Airbat\end{tabular}                            & \begin{tabular}[c]{@{}l@{}}Surfacesbois,\\ Airbat\end{tabular}                                                                         \\ \hline               
\end{tabular}
\end{table}

In table \ref{swissmuncontinuous} the multi-stage combination of the Neural Gas, Expectation Maximisation and HillClimbing algorithms provides the lowest sample size of 110 for both the atomic and continuous methods in an total training time of 95.33 seconds. This compares favourably with the lowest sample size of 120 and total training time of 1,905.82 seconds attained by the simulated annealing algorithm. The longest total training time of 547.11 seconds was for the multi-stage combination of the self-organising maps, expectation maximisation and hill climbing algorithms. 

\begin{table}[H]
\tiny
\centering
\caption{\textbf{Comparison of Sample Size, Time and Total Execution Time for the various single stage or multi-stage combinations for the Swiss Municipalities Dataset experiment}}
\label{swissmuncontinuous}
\begin{tabular}{|l|l|l|l|l|l|}
\hline
\textbf{Stage} &
  \textbf{Algorithm} &
  \textbf{Sample Size} &
  \textbf{Time (Seconds)} &
  \textbf{\begin{tabular}[c]{@{}l@{}}Total Execution \\ Time (Seconds)\end{tabular}} &
  \textbf{\begin{tabular}[c]{@{}l@{}}Total Training \\ Time (seconds)\end{tabular}} \\ \hline
\textbf{1} &
  KmeansSolution &
  271.07 &
  1.89 &
  1.89 &
  \textbf{} \\ \hline
\textbf{2} &
  \begin{tabular}[c]{@{}l@{}}K-means and \\ Hill Climbing\end{tabular} &
  229.00 &
  30.05 &
  30.05 &
  \textbf{31.94} \\ \hline
\textbf{1 and 2} &
  SOM and K-means &
  150.58 &
  4.03 &
  96.55 &
  \textbf{} \\ \hline
\textbf{3} &
  HillClimbing &
  125.31 &
  139.69 &
  139.69 &
  \textbf{236.24} \\ \hline
\textbf{1 and 2} &
  NG and K-means &
  229.10 &
  3.30 &
  66.10 &
  \textbf{} \\ \hline
\textbf{3} &
  HillClimbing &
  189.00 &
  26.53 &
  26.53 &
  \textbf{92.63} \\ \hline
\textbf{1} &
  \begin{tabular}[c]{@{}l@{}}Expectation \\ Maximisation\end{tabular} &
  174.01 &
  2.02 &
  2.02 &
  \textbf{} \\ \hline
\textbf{2} &
  HillClimbing &
  132.66 &
  131.44 &
  131.44 &
  \textbf{133.46} \\ \hline
\textbf{1 and 2} &
  SOM and EM &
  165.35 &
  14.12 &
  342.86 &
  \textbf{} \\ \hline
\textbf{3} &
  HillClimbing &
  124.48 &
  204.25 &
  204.25 &
  {\color[HTML]{FE0000} \textbf{547.11}} \\ \hline
\textbf{1 and 2} &
  NG and EM &
  124.96 &
  2.90 &
  56.72 &
  \textbf{} \\ \hline
\textbf{3} &
  HillClimbing &
  {\color[HTML]{32CB00} \textbf{110.00}} &
  38.61 &
  38.61 &
  \textbf{95.33} \\ \hline
\textbf{1} &
  Fuzzy Clustering &
  144.97 &
  0.47 &
  8.55 &
  \textbf{} \\ \hline
\textbf{2} &
  HillClimbing &
  125.17 &
  49.88 &
  49.88 &
  \textbf{58.43} \\ \hline
\textbf{1 and 2} &
  SOM and FC &
  129.64 &
  4.60 &
  94.87 &
  \textbf{} \\ \hline
\textbf{3} &
  HillClimbing &
  118.09 &
  130.38 &
  130.38 &
  \textbf{225.25} \\ \hline
\textbf{1 and 2} &
  NG and FC &
  138.23 &
  1.76 &
  35.85 &
  \textbf{} \\ \hline
\textbf{3} &
  HillClimbing &
  123.95 &
  67.91 &
  67.91 &
  \textbf{103.76} \\ \hline
\end{tabular}
\end{table}

The number of iterations for the self-organising maps algorithm in the  self-organising maps, expectation maximisation and hill climbing combination is 5,314 and significantly higher than the number of iterations for the self-organising maps algorithm in the other combinations. This (along with the training process) would contribute to explaining the longer total training time for the this multi-stage combination. 

\begin{figure}[H]
\centering 
\includegraphics[]{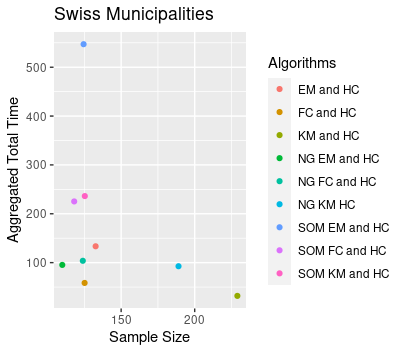}
\caption{\textbf{Plot of Aggregated Total Time versus Sample Size by Algorithm combination for the Swiss Municipalities Dataset}}
\label{ContinuousSwissMunSampleTimes}
\end{figure}

Figure \ref{ContinuousSwissMunSampleTimes} demonstrates that the NG, EM and hill climbing algorithm attains the smallest sample size in a relatively low time. A number of combinations return similar sample sizes but with varying times. The KM and hill climbing combination returns the largest sample size in the least time. 

Table \ref{SwissContinuousfine} contains the the fine-tuned hyperparameters for the various single-stage or multi-stage algorithm combinations for the Swiss Municipalities dataset.

\begin{table}[H]
\caption{\textbf{Summary of the fine-tuned hyperparameters for the various single-stage or multi-stage algorithm combinations for the Swiss Municipalities Dataset}}
\label{SwissContinuousfine}
\tiny
\centering
\begin{tabular}{|l|l|l|l|l|}
\hline
                          & \textbf{Iterations}  & \textbf{Alpha High} & \textbf{Alpha Low}     & \textbf{Radius}       \\ \hline
\textbf{SOM and K-means}   & 146                  & 0.955179772565607   & 0.097093850745587      & 0.648499123915099     \\ \hline
\textbf{}                 & \textbf{Lambda High} & \textbf{Lambda Low} & \textbf{Stepsize High} & \textbf{Stepsize Low} \\ \hline
\textbf{NG and K-means}    & 1.92036896656643     & 0.408302785790299   & 0.483817725175832      & 0.016913660635487     \\ \hline
\textbf{}                 & \textbf{Iterations}  & \textbf{Alpha High} & \textbf{Alpha Low}     & \textbf{Radius}       \\ \hline
\textbf{SOM and EM}       & 5,314                 & 0.860526216045479   & 0.036380460266579      & 0.266383153636882     \\ \hline
\textbf{}                 & \textbf{Lambda High} & \textbf{Lambda Low} & \textbf{Stepsize High} & \textbf{Stepsize Low} \\ \hline
\textbf{NG and EM}        & 1.02031964830343     & 0.965802849282413   & 0.49419928787723       & 0.046602493289255     \\ \hline
\textbf{}                 & \textbf{m}           &                     &                        &                       \\ \hline
\textbf{Fuzzy Clustering} & 2                    &                     &                        &                       \\ \hline
\textbf{}                 & \textbf{Iterations}  & \textbf{Alpha High} & \textbf{Alpha Low}     & \textbf{Radius}       \\ \hline
\textbf{SOM and FC}    & 361                  & 0.463767154766247   & 0.003181504371105      & 0.00137499391567      \\ \hline
\textbf{}                 & \textbf{Lambda High} & \textbf{Lambda Low} & \textbf{Stepsize High} & \textbf{Stepsize Low} \\ \hline
\textbf{NG and FC}     & 21.617934961514      & 0.764578349819259   & 0.133934543593412      & 0.080908668275684     \\ \hline
\end{tabular}
\end{table}

\subsection{American Community Survey, 2015}

A summary of the target and auxiliary variables for the American Survey 2015 dataset is provided in table \ref{ACSContinuous3a} below. 

\begin{table}[H]
\caption{\textbf{Summary of the Target and Auxiliary variables, number of records and number of domains for the American Survey 2015 dataset}}
\label{ACSContinuous3a}
\tiny
\centering
\begin{tabular}{|l|l|l|l|l|}
\hline
\textbf{Dataset}                                                              & \textbf{Target Variables}                                                                  & \textbf{Auxiliary Variables}                                                               & \textbf{\begin{tabular}[c]{@{}l@{}}Number of \\ Records\end{tabular}}
& \textbf{\begin{tabular}[c]{@{}l@{}}Number of \\ Domains\end{tabular}}\\
 \hline
\textbf{\begin{tabular}[c]{@{}l@{}}American Community \\ Survey, 2015\end{tabular}}    & \begin{tabular}[c]{@{}l@{}}HINCP, VALP, \\ SMOCP, INSP\end{tabular}                        & \begin{tabular}[c]{@{}l@{}}BLD, TEN, WKEXREL, \\ WORKSTAT, HFL, YBL\end{tabular}           & 619,747  & 51                                                              \\ \hline
\end{tabular}
\end{table}

The results of the various single stage or multi-stage algorithm combinations for the American Community Survey, 2015 dataset experiment are presented in table \ref{ACSContinuous3}. Note that for the K-means Hartigan-Wong method the R version of the algorithm encountered difficulties and failed to converge for either 100 (initially) or 1,000 iterations . We used the \citep{macqueen1967some} method instead- moving from 100 to 1,000 iterations in order to attain convergence. The minimum sample size attained was 2,753.55 with a total training time of 41,740.96 seconds for the staged combination of the neural gas, fuzzy clustering and hill climbing algorithms. This compares favourably with the lowest sample size of 3,915.48 and total training time of 169,115.92 seconds attained by the simulated annealing algorithm in table \ref{Continuousresults}. The longest total training time was 78,344.99 seconds for the K-means and hill climbing combination. 

\begin{table}[H]
\caption{\textbf{Comparison of Sample Size, Time and Total Execution Time for the various single stage or multi-stage algorithm combinations for the American Community Survey, 2015 Dataset}}
\label{ACSContinuous3}
\tiny
\centering
\begin{tabular}{|l|l|l|l|l|l|l|}
\hline
\textbf{Stage} &
  \textbf{Algorithm} &
  \textbf{Sample Size} &
  \textbf{Time (Seconds)} &
  \textbf{\begin{tabular}[c]{@{}l@{}}Total Execution \\ Time (Seconds)\end{tabular}} &
  \textbf{\begin{tabular}[c]{@{}l@{}}Aggregate Total \\ Time (seconds)\end{tabular}} &
  \textbf{Note} \\ \hline
\textbf{1} &
  KmeansSolution &
  8,864.89 &
  109.72 &
  109.72 &
   &
   \\ \hline
\textbf{2} &
  HillClimbing &
  3,015.11 &
  78,235.27 &
  78,235.27 &
  {\color[HTML]{F56B00} \textbf{78,344.99}} &
   \\ \hline
\textbf{1 and 2} &
  SOM and K-means &
  3,819.99 &
  313.32 &
  8,322.50 &
   &
   \\ \hline
\textbf{3} &
  HillClimbing &
  3,310.36 &
  9,882.47 &
  9,882.47 &
  18,204.97 &
   \\ \hline
\textbf{1 and 2} &
  NG and K-means &
  3,168.46 &
  657.87 &
  13,449.13 &
   &
  \begin{tabular}[c]{@{}l@{}}*MCQUEEN method\\ 1,000 iterations\end{tabular} \\ \hline
\textbf{3} &
  HillClimbing &
  2,802.93 &
  6,655.40 &
  6,655.40 &
  20,104.53 &
   \\ \hline
\textbf{1} &
  \begin{tabular}[c]{@{}l@{}}Expectation \\ Maximisation\end{tabular} &
  5,520.72 &
  141.66 &
  141.66 &
   &
   \\ \hline
\textbf{2} &
  HillClimbing &
  3,443.64 &
  46,108.23 &
  46,108.23 &
  46,249.89 &
   \\ \hline
\textbf{1 and 2} &
  SOM and EM &
  4,441.14 &
  287.36 &
  9,301.32 &
   &
   \\ \hline
\textbf{3} &
  HillClimbing &
  3,350.65 &
  24,693.78 &
  24,693.78 &
  33,995.10 &
   \\ \hline
\textbf{1 and 2} &
  NG and EM &
  3,137.56 &
  670.08 &
  13,658.88 &
   &
   \\ \hline
\textbf{3} &
  HillClimbing &
  2,808.22 &
  9,281.39 &
  9,281.39 &
  22,940.27 &
   \\ \hline
\textbf{1} &
  Fuzzy Clustering &
  5,775.91 &
  24.97 &
  2,090.77 &
   &
   \\ \hline
\textbf{2} &
  HillClimbing &
  2,825.38 &
  51,268.38 &
  51,268.38 &
  53,359.15 &
   \\ \hline
\textbf{1 and 2} &
  SOM and FC &
  3,798.98 &
  260.48 &
  7,904.15 &
   &
   \\ \hline
\textbf{3} &
  HillClimbing &
  3,193.08 &
  12,691.15 &
  12,691.15 &
  20,595.30 &
   \\ \hline
\textbf{1 and 2} &
  NG and FC &
  3,650.26 &
  653.14 &
  13,108.87 &
   &
   \\ \hline
\textbf{3} &
  HillClimbing &
  {\color[HTML]{32CB00} \textbf{2,753.55}} &
  28,632.09 &
  28,632.09 &
  41,740.96 &
   \\ \hline
\end{tabular}
\end{table}

\begin{figure}[H]
\centering 
\includegraphics[]{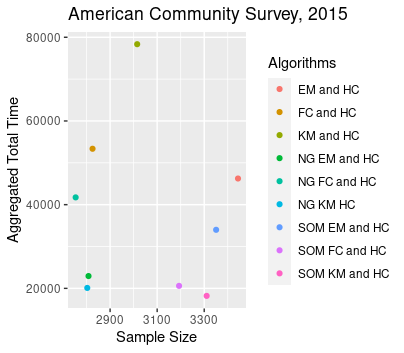}
\caption{\textbf{Plot of Aggregated Total Time versus Sample Size by Algorithm combination for the American Community Survey, 2015 Dataset}}
\label{ContinuousACSSampleTimes}
\end{figure}

Figure \ref{ContinuousACSSampleTimes} plots the aggregated total time against Sample Size by algorithm combination for the Swiss Municipalities dataset. The graph demonstrates that the EM and hill climbing algorithm attains the smallest sample size in a relatively low time. A number of combinations return similar sample sizes but with varying times. SOM, EM and hill climbing combination returns the largest sample size in the longest time. 

Table \ref{ACSContinuousFine} summarises the hyper-parameters and relevant algorithms for which the above results were attained.

\begin{table}[H]
\caption{\textbf{Summary of the fine-tuned hyperparameters for the various single-stage or multi-stage algorithm combinations for the American Community Survey, 2015 Dataset}}
\label{ACSContinuousFine}
\tiny
\centering
\begin{tabular}{|l|l|l|l|l|}
\hline
\textbf{SOM and K-means}   & \textbf{Iterations} & \textbf{Alpha High} & \textbf{Alpha Low} & \textbf{Radius}   \\ \hline
\textbf{}                 & 362                 & 0.235757044486596   & 0.012774618075309  & 0.484338735244991 \\ \hline
\textbf{NG and K-means} & \textbf{Lambda High} & \textbf{Lambda Low} & \textbf{Stepsize High} & \textbf{Stepsize Low} \\ \hline
\textbf{}                 & 22.8202557664311    & 0.916501772178628   & 0.499675245659364  & 0.008470696479746 \\ \hline
\textbf{SOM and EM}       & \textbf{Iterations} & \textbf{Alpha High} & \textbf{Alpha Low} & \textbf{Radius}   \\ \hline
\textbf{}                 & 268                 & 0.170893752936649   & 0.004722280048395  & 0.557606799123925 \\ \hline
\textbf{NG and EM}     & \textbf{Lambda High} & \textbf{Lambda Low} & \textbf{Stepsize High} & \textbf{Stepsize Low} \\ \hline
\textbf{}                 & 29.1800880556926    & 0.25795004530251    & 0.266119958357885  & 0.019287368137506 \\ \hline
\textbf{Fuzzy Clustering} & \textbf{m}          &                     &                    &                   \\ \hline
\textbf{}                 & 11                  &                     &                    &                   \\ \hline
\textbf{SOM and FC}    & \textbf{Iterations} & \textbf{Alpha High} & \textbf{Alpha Low} & \textbf{Radius}   \\ \hline
\textbf{}                 & 249                 & 0.124999162557325   & 0.052358677482419  & 0.073764869553736 \\ \hline
\textbf{NG and FC}  & \textbf{Lambda High} & \textbf{Lambda Low} & \textbf{Stepsize High} & \textbf{Stepsize Low} \\ \hline
\textbf{}                 & 5.09321314570995    & 0.984287188866795   & 0.287914934710588  & 0.00526985258796  \\ \hline
\end{tabular}
\end{table}

\subsection{Kiva Loans}

In table \ref{kivaContinuous} the lowest sample size of 1,752.82 was attained using a combination of Expectation Maximisation and Hill climbing, in a total training time of 7,428.64 seconds. This compares favourably with the lowest sample size of 3,017.79 and total training time of 4,149.06 seconds attained by the simulated annealing algorithm in table \ref{Continuousresults}. The longest total training time was 14,613.90 seconds for the self-organising maps, expectation maximisation and hill climbing combination. 

\begin{table}[H]
\caption{\textbf{Summary of the Target and Auxiliary variables, number of records and number of domains for the Kiva Loans dataset}}
\label{kivaContinuous}
\tiny
\centering
\begin{tabular}{|l|l|l|l|l|l|}
\hline
\textbf{Stage} &
  \textbf{Algorithm} &
  \textbf{Sample Size} &
  \textbf{Time (Seconds)} &
  \textbf{\begin{tabular}[c]{@{}l@{}}Total Execution \\ Time (Seconds)\end{tabular}} &
  \textbf{\begin{tabular}[c]{@{}l@{}}Aggregated Total \\ Time (seconds)\end{tabular}} \\ \hline
\textbf{1}       & KmeansSolution                                                      & 3,901.98                                 & 43.60    & 43.60    &           \\ \hline
\textbf{2}       & HillClimbing                                                        & 2,001.55                                 & 5,941.14 & 5,941.14 & 5,984.74  \\ \hline
\textbf{1 and 2} & SOM and K-means                                                      & 3,005.53                                 & 108.91   & 2,801.87 &           \\ \hline
\textbf{3}       & HillClimbing                                                        & 1,876.29                                 & 4,692.34 & 4,692.34 & 7,494.21  \\ \hline
\textbf{1 and 2} & NG and K-means                                                       & 2,636.01                                 & 160.70   & 3,269.08 &           \\ \hline
\textbf{3}       & HillClimbing                                                        & 2,043.52                                 & 3,171.44 & 3,171.44 & 6,440.52  \\ \hline
\textbf{1}       & \begin{tabular}[c]{@{}l@{}}Expectation \\ Maximisation\end{tabular} & 3,496.17                                 & 58.11    & 58.11    &           \\ \hline
\textbf{2}       & HillClimbing                                                        & {\color[HTML]{32CB00} \textbf{1,752.82}} & 7,370.53 & 7,370.53 & 7,428.64  \\ \hline
\textbf{1 and 2} & SOM and EM                                                          & 4,452.22                                 & 188.69   & 3,655.43 &           \\ \hline
\textbf{3} &
  HillClimbing &
  2,150.91 &
  10,958.47 &
  {\color[HTML]{333333} 10,958.47} &
  {\color[HTML]{F56B00} \textbf{14,613.90}} \\ \hline
\textbf{1 and 2} & NG and EM                                                           & 3,198.13                                 & 125.70   & 2,484.06 &           \\ \hline
\textbf{3}       & HillClimbing                                                        & 1,840.88                                 & 6,517.68 & 6,517.68 & 9,001.74  \\ \hline
\textbf{1}       & Fuzzy Clustering                                                    & 3,885.66                                 & 20.27    & 452.67   &           \\ \hline
\textbf{2}       & HillClimbing                                                        & 1,879.47                                 & 6,559.08 & 6,559.08 & 7,011.75  \\ \hline
\textbf{1 and 2} & SOM and FC                                                          & 2,547.45                                 & 98.68    & 2,878.54 &           \\ \hline
\textbf{3}       & HillClimbing                                                        & 1,818.98                                 & 4,501.45 & 4,501.45 & 7,379.99  \\ \hline
\textbf{1 and 2} & NG and FC                                                           & 3,351.17                                 & 93.50    & 1,875.48 &           \\ \hline
\textbf{3}       & HillClimbing                                                        & 1,869.71                                 & 9,528.57 & 9,528.57 & 11,404.05 \\ \hline
\end{tabular}
\end{table}

\begin{figure}[H]
\centering 
\includegraphics[]{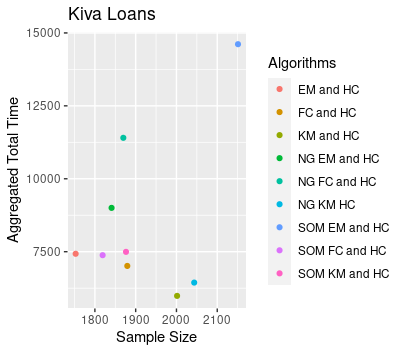}
\caption{\textbf{Plot of Aggregated Total Time versus Sample Size by Algorithm combination for the Kiva Loans Dataset}}
\label{ContinuousKivaLoansSampleTimes}
\end{figure}

Figure \ref{ContinuousKivaLoansSampleTimes} plots the aggregated total time against Sample Size by algorithm combination for the Swiss Municipalities dataset. The graph demonstrates that the EM and hill climbing algorithm attains the smallest sample size in a relatively low time. A number of combinations return similar sample sizes but with varying times. SOM, EM and hill climbing combination returns the largest sample size in the longest time. 

In table \ref{KivafineContin} we provide details of the fine-tuned hyper-parameters for the relevant algorithms.

\begin{table}[H]
\caption{\textbf{Summary of the fine-tuned hyperparameters for the various single-stage or multi-stage algorithm combinations for the American Community Survey, 2015 Dataset}}
\label{KivafineContin}
\tiny
\centering
\begin{tabular}{|l|l|l|l|l|}
\hline
\textbf{SOM and K-means}   & \textbf{Iterations}  & \textbf{Alpha High} & \textbf{Alpha Low}     & \textbf{Radius}       \\ \hline
                          & 371                  & 0.277877372112223   & 0.012899151606656      & 0.694571047057514     \\ \hline
\textbf{NG and K-means} & \textbf{Lambda High} & \textbf{Lambda Low} & \textbf{Stepsize High} & \textbf{Stepsize Low} \\ \hline
                          & 15.1897164925445     & 0.822356168348829   & 0.10500027609291       & 0.005030299354611     \\ \hline
\textbf{SOM and EM}       & \textbf{Iterations}  & \textbf{Alpha High} & \textbf{Alpha Low}     & \textbf{Radius}       \\ \hline
                          & 914                  & 0.162357069495344   & 0.040045055053234      & 0.535640629408619     \\ \hline
\textbf{NG and EM}        & \textbf{Lambda High} & \textbf{Lambda Low} & \textbf{Stepsize High} & \textbf{Stepsize Low} \\ \hline
                          & 20.9163182787887     & 0.804910189211671   & 0.183011407460269      & 0.076024879621262     \\ \hline
\textbf{Fuzzy Clustering} & \textbf{m}           & \textbf{}           & \textbf{}              & \textbf{}             \\ \hline
                          & 2                    &                     &                        &                       \\ \hline
\textbf{SOM and FC}       & \textbf{Iterations}  & \textbf{Alpha High} & \textbf{Alpha Low}     & \textbf{Radius}       \\ \hline
                          & 264                  & 0.113939729200586   & 0.00350977823364       & 0.003857035322069     \\ \hline
\textbf{NG and FC}        & \textbf{Lambda High} & \textbf{Lambda Low} & \textbf{Stepsize High} & \textbf{Stepsize Low} \\ \hline
                          & 18.9297597394552     & 0.643763823658483   & 0.136589161984179      & 0.067314184908353     \\ \hline
\end{tabular}
\end{table}

\subsection{Benchmark Results - Continuous Strata}

For comparison purposes table \ref{Continuousresults} contains the evaluation time, sample size and number of strata for the Grouping Genetic Algorithm (GGA) and Simulated Annealing Algorithm (SAA) for the above datasets.   

\begin{table}[H]
\caption{\textbf{Summary by dataset of the evaluation time, sample size and number of strata for the Grouping Genetic Algorithm (GGA) and Simulated Annealing Algorithm (SAA)}}
\label{Continuousresults}
\centering
\tiny
\begin{tabular}{|l|l|l|l|l|l|l|}
\hline
\textbf{Dataset} &
  \textbf{Sample Size} &
  \textbf{\begin{tabular}[c]{@{}l@{}}Execution \\ Time\end{tabular}} &
  \textbf{\begin{tabular}[c]{@{}l@{}}Total \\Training \\ Time\end{tabular}} &
  \textbf{Sample Size} &
  \textbf{\begin{tabular}[c]{@{}l@{}}Execution \\ Time\end{tabular}} &
  \textbf{\begin{tabular}[c]{@{}l@{}}Total\\ Training \\ Time\end{tabular}} \\ \hline
\textbf{SwissMunicipalities}                & 128.69   & 753.82    & 10,434.30  & 120.00   & 213.44    & 1,905.82   \\ \hline
\textbf{American Community Survey, 2015}    & 4,197.68 & 22,016.95 & 227,635.51 & 3915.48  & 13,351.19 & 169,115.92 \\ \hline
\textbf{Kiva Loans}                         & 3,062.33 & 3,232.78  & 48,746.61  & 3,017.79 & 300.16    & 4,149.06   \\ \hline
\end{tabular}
\end{table}

\section{Conclusions and Future work}\label{Conclusions}

The multi-stage combination of k-means type algorithms with hill climbing performed better, generally, than the grouping genetic algorithm and the simulated annealing algorithm for atomic and continuous strata. The exception being the case of atomic strata for the Swiss Municipalities dataset, where the grouping genetic algorithm performed better- although the survey designer may wish to consider solution quality in the context of training times (as with all the other experiments). The continuous strata provide lower sample samples than the atomic strata method. This is because the stratification variable is the target variable - which results in better efficiency.\\ The performance of each multi-stage combination of algorithms depends on the dataset and the auxiliary and stratification variables. Therefore there isn't one combination that suits all experiments. Nonetheless, we have provided the survey designer with a choice of combinations, which can be applied in the context expert knowledge of the particular sampling problem. For future work further combinations of the hill climbing algorithm with other clustering algorithms such as affinity propagation clustering, cobweb clustering algorithm, density-based clustering, divisive hierarchical clustering, simple featureless clustering, farthest first clustering algorithm, mean shift clustering, kernel k-means clustering, mini batch k-means clustering, k-means with automatic determination of k, partitioning around medoids, k-medians clustering, etc. may be considered. Alternative local search algorithms such as tabu search, stochastic hill climbing, local beam search or simulated annealing may also be considered for the final stage. The grouping genetic algorithm may also be applied. 

\section*{Acknowledgements}

This material is based upon work supported by the Insight Centre for Data Analytics and Science Foundation Ireland under Grant No. 12/RC/2289-P2 which is co-funded under the European Regional Development Fund. Also, this publication has emanated from research supported in part by a grant from Science Foundation Ireland under Grant number 16/RC/3918 which is co-funded under the European Regional Development Fund.


\begin{thebibliography}{}


  
  \bibitem[\protect\citeauthoryear{Asan and Ercan}{Asan and
  Ercan}{2012}]{asan2012introduction}
Asan, U. and S.~Ercan (2012).
\newblock An introduction to self-organizing maps.
\newblock In {\em Computational Intelligence Systems in Industrial
  Engineering}, pp.\  295--315. Springer.
  
  \bibitem[\protect\citeauthoryear{{\"A}yr{\"a}m{\"o} and
  K{\"a}rkk{\"a}inen}{{\"A}yr{\"a}m{\"o} and
  K{\"a}rkk{\"a}inen}{2006}]{ayramo2006introduction}
{\"A}yr{\"a}m{\"o}, S. and T.~K{\"a}rkk{\"a}inen (2006).
\newblock Introduction to partitioning-based clustering methods with a robust
  example.
\newblock {\em Reports of the Department of Mathematical Information
  Technology. Series C, Software engineering and computational
  intelligence\/}~(1/2006).

\bibitem[\protect\citeauthoryear{Ba{\c{c}}{\~a}o, Lobo, and
  Painho}{Ba{\c{c}}{\~a}o et~al.}{2005}]{baccao2005self}
Ba{\c{c}}{\~a}o, F., V.~Lobo, and M.~Painho (2005).
\newblock Self-organizing maps as substitutes for k-means clustering.
\newblock In {\em International Conference on Computational Science}, pp.\
  476--483. Springer.
  
\bibitem[\protect\citeauthoryear{Ballin and Barcaroli}{Ballin and
  Barcaroli}{2013}]{ballin2013joint}
Ballin, M. and G.~Barcaroli (2013).
\newblock Joint determination of optimal stratification and sample allocation
  using genetic algorithm.
\newblock {\em Survey Methodology\/}~{\em 39\/}(2), 369--393.

\bibitem[\protect\citeauthoryear{Ballin and Barcaroli}{Ballin and
  Barcaroli}{2020}]{ballin2020optimization}
Ballin, M. and G.~Barcaroli (2020).
\newblock Optimization of sampling strata with the SamplingStrata package.
\newblock Accessed 3 May 2021.
\newblock https://barcaroli.github.io/SamplingStrata/articles/SamplingStrata.html

\bibitem[\protect\citeauthoryear{Barcaroli et~al.}{Barcaroli
  et~al.}{2014}]{barcaroli2014samplingstrata} Barcaroli, G. (2014).
\newblock SamplingStrata: An {R} package for the optimization of stratified sampling.
\newblock {\em Journal of Statistical Software\/}~{\em 61\/}(4), 1--24.


\bibitem[\protect\citeauthoryear{Bethel}{Bethel}{1985}]{bethel1985optimum}
Bethel, J.~W. (1985).
\newblock {\em An optimum allocation algorithm for multivariate surveys}.
\newblock American Statistical Proceedings of the Survey Research Methods,
  Section, 209-212.    
  
\bibitem[\protect\citeauthoryear{Bethel}{Bethel}{1989}]{bethel1989sample}
Bethel, J. (1989).
\newblock Sample allocation in multivariate surveys.
\newblock {\em Survey methodology\/}~{\em 15\/}(1), 47\--57.

\bibitem[\protect\citeauthoryear{Bezdek, Ehrlich, and Full}{Bezdek
  et~al.}{1984}]{bezdek1984fcm}
Bezdek, J.~C., R.~Ehrlich, and W.~Full (1984).
\newblock Fcm: The fuzzy c-means clustering algorithm.
\newblock {\em Computers \& geosciences\/}~{\em 10\/}(2-3), 191--203.

\bibitem[\protect\citeauthoryear{Choudhary}{Choudhary}{2015}]{ankitchoudhary2015}
Choudhary, A. (2015, 10).
\newblock Empirical comparison of performances of k-means, k-means++, weighted
  k-means and hartigan and wong k-means clustering algorithms.

\bibitem[\protect\citeauthoryear{Chromy}{Chromy}{1987}]{chromy1987design}
Chromy, J.~R. (1987).
\newblock Design optimization with multiple objectives.
\newblock {\em Proceedings of the Survey Research Methods Section\/}.

\bibitem[\protect\citeauthoryear{Dempster, Laird, and Rubin}{Dempster
  et~al.}{1977}]{dempster1977maximum}
Dempster, A.~P., N.~M. Laird, and D.~B. Rubin (1977).
\newblock Maximum likelihood from incomplete data via the em algorithm.
\newblock {\em Journal of the Royal Statistical Society: Series B
  (Methodological)\/}~{\em 39\/}(1), 1--22.

\bibitem[\protect\citeauthoryear{Dimitriadou, Hornik, and Hornik}{Dimitriadou
  et~al.}{2020}]{dimitriadou2020package}
Dimitriadou, E., K.~Hornik, and M.~K. Hornik (2020).
\newblock Package ‘cclust’.

\bibitem[\protect\citeauthoryear{Dirgov{\'a}~Lupt{\'a}kov{\'a}, {\v{S}}imon,
  Huraj, and Posp{\'\i}chal}{Dirgov{\'a}~Lupt{\'a}kov{\'a}
  et~al.}{2016}]{dirgova2016neural}
Dirgov{\'a}~Lupt{\'a}kov{\'a}, I., M.~{\v{S}}imon, L.~Huraj, and
  J.~Posp{\'\i}chal (2016).
\newblock Neural gas clustering adapted for given size of clusters.
\newblock {\em Mathematical Problems in Engineering\/}~{\em 2016}.


\bibitem[\protect\citeauthoryear{Drineas, Frieze, Kannan, Vempala, and
  Vinay}{Drineas et~al.}{2004}]{drineas2004clustering}
Drineas, P., A.~Frieze, R.~Kannan, S.~Vempala, and V.~Vinay (2004).
\newblock Clustering large graphs via the singular value decomposition.
\newblock {\em Machine learning\/}~{\em 56\/}(1), 9--33.

\bibitem[\protect\citeauthoryear{Forgy}{Forgy}{1965}]{forgy1965cluster}
Forgy, E.~W. (1965).
\newblock Cluster analysis of multivariate data: efficiency versus
  interpretability of classifications.
\newblock {\em biometrics\/}~{\em 21}, 768--769.

\bibitem[\protect\citeauthoryear{Friedman, Hastie, Tibshirani, et~al.}{Friedman
  et~al.}{2001}]{friedman2001elements}
Friedman, J., T.~Hastie, R.~Tibshirani, et~al. (2001).
\newblock {\em The elements of statistical learning}, Volume~1.
\newblock Springer series in statistics New York.

\bibitem[\protect\citeauthoryear{Hartigan}{Hartigan}{1979}]{hartigan1979k}
Hartigan, J.~A. (1979).
\newblock A k-means clustering algorithm: Algorithm as 136.
\newblock {\em Appl. Stat.\/}~{\em 28}, 126--130.

  
  
\bibitem[\protect\citeauthoryear{Jain}{Jain}{2010}]{jain2010data}
Jain, A.~K. (2010).
\newblock Data clustering: 50 years beyond k-means.
\newblock {\em Pattern recognition letters\/}~{\em 31\/}(8).


\bibitem[\protect\citeauthoryear{Klawonn}{Klawonn}{2004}]{klawonn2004fuzzy}
Klawonn, F. (2004).
\newblock Fuzzy clustering: Insights and a new approach.
\newblock {\em Mathware \& soft computing. 2004 Vol. 11 Num. 3\/}.


\bibitem[\protect\citeauthoryear{DZone.com}{}{}]{k-meansSOMIntroduction}
K-{{Means}} and {{SOM}}: {{Introduction}} to {{Popular Clustering Algorithms}}
  - {{DZone AI}}.
\newblock
  https://dzone.com/articles/k-means-and-som-gentle-introduction-to-worlds-most.
  \newblock Accessed 25th May 2021.
  
  
  \bibitem[\protect\citeauthoryear{Kohonen}{Kohonen}{1990}]{kohonen1990self}
Kohonen, T. (1990).
\newblock The self-organizing map.
\newblock {\em Proceedings of the IEEE\/}~{\em 78\/}(9), 1464--1480.

\bibitem[\protect\citeauthoryear{Li}{Li}{2014}]{Li2014Smile}
Li, H. (2014).
\newblock Smile.
\newblock \url{https://haifengl.github.io}.
\newblock Accessed 08th June 2021.

\bibitem[\protect\citeauthoryear{Lin and Kernighan}{Lin and
  Kernighan}{1973}]{lin1973effective}
Lin, S. and B.~W. Kernighan (1973).
\newblock An effective heuristic algorithm for the traveling-salesman problem.
\newblock {\em Operations research\/}~{\em 21\/}(2), 498--516.

\bibitem[\protect\citeauthoryear{Lisic, Sang, Zhu, and Zimmer}{Lisic
  et~al.}{2018}]{lisic2018optimal}
Lisic, J., H.~Sang, Z.~Zhu, and S.~Zimmer (2018).
\newblock Optimal stratification and allocation for the june agricultural
  survey.
\newblock {\em Journal of Official Statistics\/}~{\em 34\/}(1), 121--148.

\bibitem[\protect\citeauthoryear{Lloyd}{Lloyd}{1982}]{lloyd1982least}
Lloyd, S. (1982).
\newblock Least squares quantization in pcm.
\newblock {\em IEEE transactions on information theory\/}~{\em 28\/}(2),
  129--137.
  

\bibitem[\protect\citeauthoryear{MacQueen et~al.}{MacQueen
  et~al.}{1967}]{macqueen1967some}
MacQueen, J. et~al. (1967).
\newblock Some methods for classification and analysis of multivariate
  observations.
\newblock In {\em Proceedings of the fifth Berkeley symposium on mathematical
  statistics and probability}, Volume~1, pp.\  281--297. Oakland, CA, USA.

\bibitem[\protect\citeauthoryear{Martinetz, Schulten, et~al.}{Martinetz
  et~al.}{1991}]{martinetz1991neural}
Martinetz, T., K.~Schulten, et~al. (1991).
\newblock A "neural-gas" network learns topologies.

\bibitem[\protect\citeauthoryear{Meil{\u{a}}}{Meil{\u{a}}}{2006}]{meilua2006uniqueness}
Meil{\u{a}}, M. (2006).
\newblock The uniqueness of a good optimum for k-means.
\newblock In {\em Proceedings of the 23rd international conference on Machine
  learning}, pp.\  625--632.

\bibitem[\protect\citeauthoryear{Meyer, Dimitriadou, Hornik, Weingessel,
  Leisch, Chang, Lin, and Meyer}{Meyer et~al.}{2019}]{meyer2019package}
Meyer, D., E.~Dimitriadou, K.~Hornik, A.~Weingessel, F.~Leisch, C.-C. Chang,
  C.-C. Lin, and M.~D. Meyer (2019).
\newblock Package ‘e1071’.
\newblock {\em The R Journal\/}.

\bibitem[\protect\citeauthoryear{Michalewicz and Fogel}{Michalewicz and
  Fogel}{2013}]{michalewicz2013solve}
Michalewicz, Z. and D.~B. Fogel (2013).
\newblock {\em How to solve it: modern heuristics}.
\newblock Springer Science \& Business Media.

\bibitem[\protect\citeauthoryear{Morissette and Chartier}{Morissette and
  Chartier}{2013}]{morissette2013k}
Morissette, L. and S.~Chartier (2013).
\newblock The k-means clustering technique: General considerations and
  implementation in mathematica.
\newblock {\em Tutorials in Quantitative Methods for Psychology\/}~{\em
  9\/}(1), 15--24.

\bibitem[\protect\citeauthoryear{Nayak, Naik, and Behera}{Nayak
  et~al.}{2015}]{nayak2015fuzzy}
Nayak, J., B.~Naik, and H.~Behera (2015).
\newblock Fuzzy c-means (fcm) clustering algorithm: a decade review from 2000
  to 2014.
\newblock {\em Computational intelligence in data mining-volume 2\/}, 133--149.

\bibitem[\protect\citeauthoryear{O'Luing, Prestwich, and Tarim}{O'Luing et~al.}{2019}]{oluing2019grouping}
O'Luing, M., S.~Prestwich, and S.~Armagan Tarim (2019).
\newblock A grouping genetic algorithm for joint stratification and sample allocation designs.
\newblock {\em Survey Methodology\/}~{\em 45\/}(3), 513--531.

\bibitem[\protect\citeauthoryear{O'Luing, Prestwich, and Tarim}{O'Luing
  et~al.}{2020}]{o2020simulated}
O'Luing, M., S.~Prestwich, and S.~A. Tarim (2020).
\newblock A simulated annealing algorithm for joint stratification and sample
  allocation designs.
\newblock {\em arXiv preprint arXiv:2011.13006\/}.

\bibitem[\protect\citeauthoryear{P{\'e}rez-Ortega, Almanza-Ortega,
  Vega-Villalobos, Pazos-Rangel, Zavala-D{\'\i}az, and
  Mart{\'\i}nez-Rebollar}{P{\'e}rez-Ortega et~al.}{2019}]{perez2019k}
P{\'e}rez-Ortega, J., N.~N. Almanza-Ortega, A.~Vega-Villalobos,
  R.~Pazos-Rangel, C.~Zavala-D{\'\i}az, and A.~Mart{\'\i}nez-Rebollar (2019).
\newblock The k-means algorithm evolution.
\newblock In {\em Introduction to Data Science and Machine Learning}.
  IntechOpen.
  
  \bibitem[\protect\citeauthoryear{{R Core Team}}{{R Core Team}}{2021}]{R2021}
{R Core Team} (2021).
\newblock {\em R: A Language and Environment for Statistical Computing}.
\newblock Vienna, Austria: R Foundation for Statistical Computing.

\bibitem[\protect\citeauthoryear{Scrucca, Fop, Murphy, and Raftery}{Scrucca
  et~al.}{2016}]{mclust2016}
Scrucca, L., M.~Fop, T.~B. Murphy, and A.~E. Raftery (2016).
\newblock {mclust} 5: clustering, classification and density estimation using
  {G}aussian finite mixture models.
\newblock {\em The {R} Journal\/}~{\em 8\/}(1), 289--317.

\bibitem[\protect\citeauthoryear{S{\"o}rensen and Glover}{S{\"o}rensen and
  Glover}{2013}]{sorensen2013metaheuristics}
S{\"o}rensen, K. and F.~Glover (2013).
\newblock Metaheuristics.
\newblock {\em Encyclopedia of operations research and management
  science\/}~{\em 62}, 960--970.

\bibitem[\protect\citeauthoryear{Silviu}{Silviu}{2001}]{silviu2001fuzzy}
Silviu, B. (2001).
\newblock Fuzzy clustering.
\newblock {\em Babes-Bolyai University\/}.

\bibitem[\protect\citeauthoryear{Vesanto and Alhoniemi}{Vesanto and
  Alhoniemi}{2000}]{vesanto2000clustering}
Vesanto, J. and E.~Alhoniemi (2000).
\newblock Clustering of the self-organizing map.
\newblock {\em IEEE Transactions on neural networks\/}~{\em 11\/}(3), 586--600.  

\bibitem[\protect\citeauthoryear{Wehrens, Buydens, et~al.}{Wehrens
  et~al.}{}]{wehrens2018self}
Wehrens, R., L.~M. Buydens, et~al.
\newblock Flexible self-organizing maps in kohonen 3.0.
\newblock {\em Journal of Statistical Software\/}.


\end{thebibliography}
\end{document}